\documentclass[10pt,twocolumn,letterpaper]{article}

\usepackage[pagenumbers]{cvpr} 

\usepackage{graphicx}
\usepackage{amsmath}
\usepackage{amssymb}
\usepackage{booktabs}

\usepackage{multirow}
\usepackage{algorithm}
\usepackage{algorithmic}
\usepackage[pagebackref,breaklinks,colorlinks]{hyperref}

\usepackage[capitalize]{cleveref}
\crefname{section}{Sec.}{Secs.}
\Crefname{section}{Section}{Sections}
\Crefname{table}{Table}{Tables}
\crefname{table}{Tab.}{Tabs.}

\def\cvprPaperID{8540} 
\def\confName{CVPR}
\def\confYear{2023}

\begin{document}


\title{Leveraging Angular Information Between Feature and Classifier\\ for Long-tailed Learning: A Prediction Reformulation Approach}

\author{Haoxuan Wang\\
Shanghai Jiao Tong University\\
Shanghai, China\\
{\tt\small hatchet25@sjtu.edu.cn}
\and
Junchi Yan\\
Shanghai Jiao Tong University\\
Shanghai, China\\
{\tt\small yanjunchi@sjtu.edu.cn}
}
\maketitle

\begin{abstract}
   Deep neural networks still struggle on long-tailed image datasets, and one of the reasons is that the imbalance of training data across categories leads to the imbalance of trained model parameters. Motivated by the empirical findings that trained classifiers yield larger weight norms in head classes, we propose to reformulate the recognition probabilities through included angles without re-balancing the classifier weights. Specifically, we calculate the angles between the data feature and the class-wise classifier weights to obtain angle-based prediction results. Inspired by the performance improvement of the predictive form reformulation and the outstanding performance of the widely used two-stage learning framework, we explore the different properties of this angular prediction and propose novel modules to improve the performance of different components in the framework. Our method is able to obtain the best performance among peer methods without pretraining on CIFAR10/100-LT and ImageNet-LT. Source code will be made publicly available.
\end{abstract}




\section{Introduction}
\label{sec:intro}
Long-tailed (LT) distribution is a prevalent phenomenon in machine learning and computer vision, occurring in various aspects such as sample distribution~\cite{border_smote} and feature distribution~\cite{glt}. One of the most typical settings is the imbalance in class sample numbers, introduced by the different sample collection cost and difficulties \cite{ina2018}. With the increasing amount of data collected and the growing demand of recognizing objects as refined as possible, the need for learning on long-tailed distributions is becoming inevitable. Meanwhile, many existing popular image datasets e.g. ImageNet \cite{imagenet} are often constructed with a balanced distribution. Methods verified on these datasets may face unintended difficulties when applied to imbalanced datasets. Long-tailed recognition (LTR), in this way, is introduced as a general and challenging task.

\begin{figure}[t]
  \centering
   \includegraphics[width=0.95\linewidth]{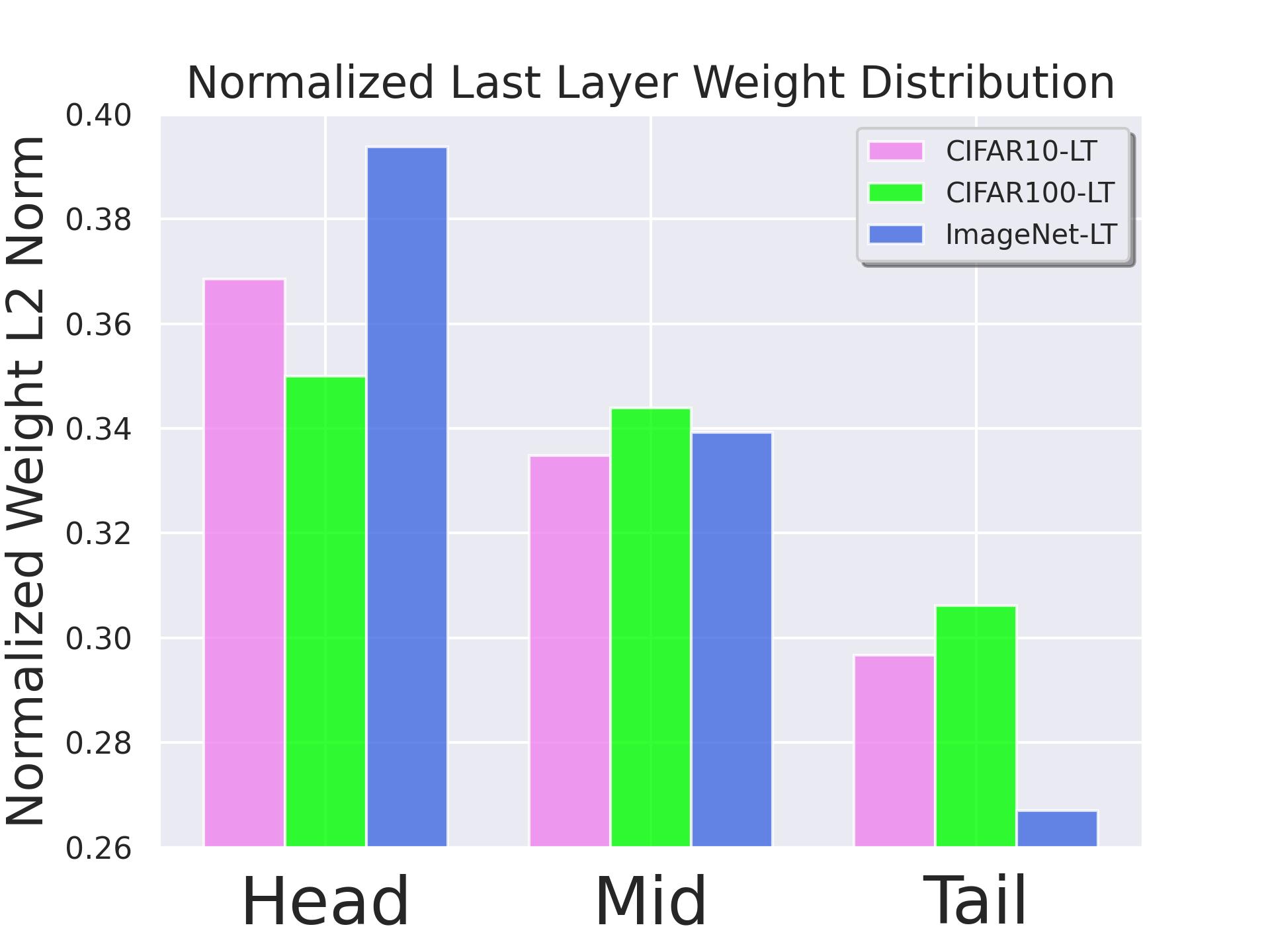}\vspace{-5pt}
   \caption{An example of model weight distribution learned on long-tailed data. The weights are averaged per class and normalized into the range of [0, 1]. We show that there is a general problem for CNN models when trained on long-tailed datasets: The weight norms of the classifier are also imbalanced, with the head classes' weights drastically larger than those of the tail classes.}
   \label{fig:weight_imb}
\end{figure}

The major distinction of long-tailed recognition with common classification tasks is the difference in class sample numbers, where some classes (tail classes) have substantially fewer samples than others (head classes). This difference leads to the critical problem in long-tailed recognition: poor performance in tail classes. Thus data augmentation methods \cite{metasaug,mixup} were proposed to expand the tail class sample numbers. However, the lack of data is not the only reason. Some works \cite{LTRweightbalancing} assume that the long-tailed distribution tempt the algorithm to focus more on head class information, aggravating the imbalance. There are also works e.g. \cite{trans_learning} that in contrast, credit the head class learning for providing robust representation abilities, and further transfer knowledge from head classes to tail classes. While the motivations are diverse, we intend to go into the specific detailed effects that the class imbalance has on the trained models. Thus we raise the question: \textit{How does the class imbalance effect the long-tailed recognition model learned?}

It is shown in \cite{bgs} that the weight of the classification layer is imbalanced under the setting of long-tailed large vocabulary object detection. We also show that the classifier weight norms are imbalanced in long-tailed recognition, verified in Fig.~\ref{fig:weight_imb}, where the training backbone used for CIFAR10-LT and CIFAR100-LT \cite{ldam} is ResNet32, and the backbone for ImageNet-LT is ResNet50. \cite{bgs} solves this issue by proposing a balanced group softmax module and using group-wise training, so that the head and tail classes are both sufficiently trained. \cite{LTRweightbalancing} uses weight decay to penalize larger weights more and uses the MaxNorm constraint \cite{maxnorm} to encourage growing small weights. Orthogonal from these methods, we resolve this problem without re-balancing the weights of the classifier. Instead, we directly leverage the trained weights by reformulating the prediction form into the angles between features and classifier weights. We further explore different properties of this angular information and utilize them in the two-stage learning framework, proposing a novel perspective for LTR.

\begin{figure}[t!]
  \centering
   \includegraphics[width=0.95\linewidth]{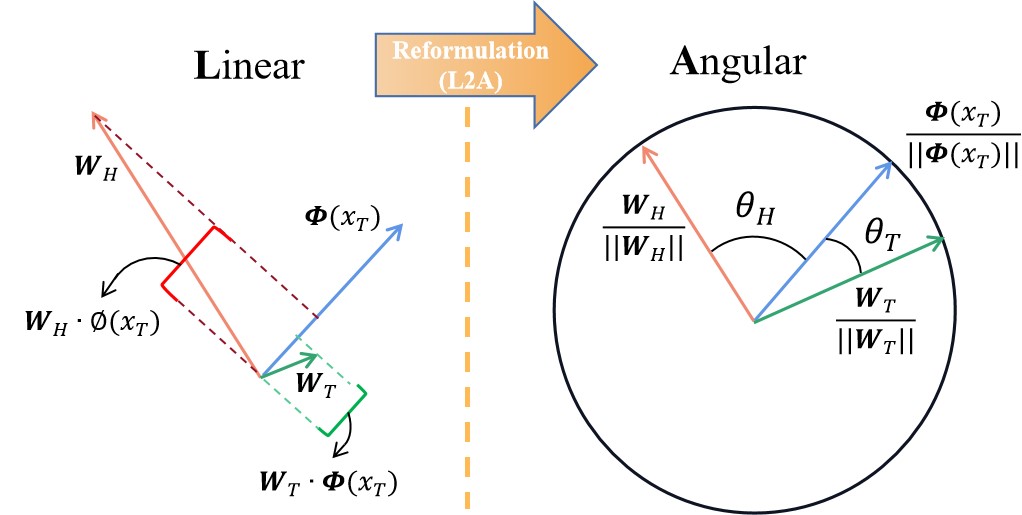}
   \caption{Rationale illustration for the prediction reformulation. $\boldsymbol{\phi}(\mathbf{x}_T)$ represents the feature output for tail class $T$'s image $x$. $\mathbf{W}_{H}$ is the classifier weight vector for head class $H$, $\mathbf{W}_{T}$ is for tail class $T$. The linear prediction compares the projection length of $\mathbf{W}_{H}$ and $\mathbf{W}_{T}$ onto $\boldsymbol{\phi}(\mathbf{x}_T)$, predicting the sample as head class. Alternately, the angular prediction classifies $x$ as belonging to $T$ by comparing the angles between the feature vector and the classifier weight vectors. This predictive form avoid the inclusion of weight imbalance by normalizing the feature vector and the classifier weight.}
   \vspace{-5pt}
   \label{fig:ang_decrib}
\end{figure}

Angular information has been used in various aspects of deep learning. Angle cosine values were integrated into the softmax loss to increase the correct categories' decision margins \cite{largemargin}. Angular distances such as the distances between feature maps \cite{decoupledNet} in CNNs were used to measure semantic differences between samples. Angular visual hardness \cite{avh} measures the angle between the feature vector and the classifier weight of the correct category, and aims to measure sample hardness. Different from these works, we intend to calculate the angles between the feature vector and the classifier's weights to form angular predictions, as a replacement or assistant to the original output. 

Our basic approach is a natural change in the predictive form without additional model training. We refer to the original output as linear prediction, which is formulated as the inner product between the feature vector of image $x$ and the classifier weight for class $c$: $P_{c}^{L} = \boldsymbol{\phi}(x) \cdot \mathbf{W}_{c}$, where $\boldsymbol{\phi}(x)$ denotes the feature vector output before the last layer, $\mathbf{W}_{c}$ denotes the layer weight for the $c$th class. We use a different form of the prediction output by directly using $\mathbf{W}_{c}$ and $\boldsymbol{\phi}(x)$. We further exploit the different properties of this prediction to utilize the long-tailed learning process. 

\textbf{The contributions of this paper include}:

1) To address the unwanted classifier weight imbalance in LTR as recently observed~\cite{bgs}, for the first time to our best knowledge in LTR, we propose introducing the angular predictions between the last layer feature vectors and the classifier weights to construct prediction logits.

2) Technically, we exploit two properties of angular predictions in long-tailed learning and propose a novel angular version of the two stage learning framework~\cite{mislas}. We show that our three new methods can enhance the performance of the original approaches, save the number of parameters to be learned, and used as simple plugin modules.

3) Results on CIFAR10/100-LT and ImageNet-LT show that our method can achieve the best performance among peer methods~\cite{mislas} without pretraining, while slightly under-perform the pretraining-based contrastive methods~\cite{bcl} and ensembling methods~\cite{ncl}. We further extend the adoption of angular information as a plug-in, to the embodiment with vision transformers and data pruning and improve their performance notably on LT data.
\section{Related Works}
There is a rich literature in long-tailed recognition, and we  discuss the most related works to better present the background of our idea and the differences with other methods.

\textbf{Designing the Classifier.}
The most common approach used for visual classification is the linear classifier, calculating the inner product of feature representation and classifier weight. However, the linear classifier is easily biased towards the head classes and cosine classifier \cite{embedaugment,ad_robust} is proposed, where both the feature and the classifier weight are normalized. A temperature factor is also applied on the normalized inner product, but need to be chosen carefully to avoid performance degradation \cite{identifyingAC}. $\tau$-normalized classifier~\cite{decoupling} only normalizes the classifier weight using different normalization intensity that are learned via class-balanced sampling. Causal classifier \cite{tde} turn to causal inference, aiming to keep the advantageous factors that stabilizes gradients and accelerates training, and removing harmful factors that aggravate long-tailed bias. Our work is different from these methods in three aspects: First, we do not strictly design a classifier which requires parameter learning. Instead, we directly employ the learned parameters from the linear classifier. Second, we use angles as classification basis instead of cosine classifier's trigonometric values. Third, \cite{embedaugment} and \cite{ad_robust} only focus on face recognition and adversarial robustness, not for long-tailed recognition.

\textbf{Preprocessing Data.}
To deal with the imbalance in data, the intuitive idea is to re-balance the data. Over-sampling \cite{learning-imb, oversample1, oversample2} the instances in tail classes or under-sampling \cite{undersample1, undersample2} the instances in head classes are two common ways. However, over-sampling may lead to overfit in tail classes since the same instances are copied again and again; Under-sampling may cause poorly learned head class information and much valuable information are lost. Apart from sampling, data-augmentation is another direction for re-balancing the tail classes. Both image-level augmentation and feature-level augmentation are proposed. Image-level augmentation such as Remix \cite{remix} improves Mixup \cite{mixup} by assigning assigning higher weights to the tail class labels. Instance-level augmentation such as SMOTE \cite{smote} finds the nearest neighbor of each minority class instance and uses a random linear combination of them as a new sample of that class. \cite{mislas} found that Mixup is able to ease the weight imbalance in the classifier, however, Fig. \ref{fig:weight_imb} shows that Mixup cannot fully resolve this problem. In our work, we only use over-sampling and simple Mixup for fair comparison with baselines.

\textbf{Decoupling Representation and Classifier.}
Recent studies like BBN \cite{bbn} and network decoupling \cite{decoupling} show that separating the training process into two stages is beneficial for LTR. They propose to firstly train the whole neural network with long-tailed data to obtain high quality feature representations. Then freeze the feature extraction layers and only finetune the classification layer using re-sampled balanced data. MiSLAS \cite{mislas} is a solid work that provides a rigorous setting, and enhances the performance using a class re-balanced version of label-aware smoothing. Two-stage learning is also expanded to visual-language areas \cite{vl-ltr}, where the improvement mainly comes from better learned features. Our work propose an angular version of the two stage learning framework \cite{mislas} by integrating angular information into different modules. Whereas decoupling methods achieves significant performance, they are against the end-to-end training preference in deep learning. Thus we discuss the performance in both stages.


\section{Understanding the Effect of Angular Information in Long-tailed Learning}
We discuss the properties of angular predictions. Motivated by these findings, we develop different methods for using angular information to tackle long-tailed recognition. We base our observations on the two-stage learning framework \cite{mislas}, and analyze the two stages independently as well as jointly. For all the settings that we discuss, the classes are sorted in decreasing order of sample numbers (the larger the class index, the fewer samples it contains). 

\subsection{Reformulating the Model Prediction}
Motivated by Fig. \ref{fig:ang_decrib}, we believe that the widely-accepted inner product output, which we refer to as the linear prediction, is not reasonable when the classifier weight magnitudes are imbalanced. Different from the linear prediction $P_{c}^{L} = \boldsymbol{\phi}(x) \cdot \mathbf{W}_{c}$, we would like to characterize the angular prediction as the angular similarity between the feature vector and the classifier weight:
\begin{equation}
    P_{c}^{A} = \pi - \arccos\left(\frac{\boldsymbol{\phi}(x)\cdot \mathbf{W}_{c}}{\|\boldsymbol{\phi}(x)\|\|\mathbf{W}_{c}\|}\right).
\label{eq:ang}
\end{equation}

Take Fig. \ref{fig:ang_decrib} as an binary classification example, suppose the feature vector for an tail class sample $x_{T}$ is $\phi(x_{T})$, $W_{H}$ and $W_{T}$ are the weights for head class $H$ and ground-truth tail class $T$, respectively. The inner product output (linear prediction) is calculated via:
\begin{equation}
    \mathbf{W}_{H} \cdot \boldsymbol{\phi}(x_{T}) > \mathbf{W}_{T} \cdot \boldsymbol{\phi}(x_{H}), 
\end{equation}
classifying $x_{T}$ as belonging to class $H$. On the other hand, for making the correct prediction, the angular prediction is:
\begin{equation}
    \pi - \theta_{H} < \pi - \theta_{T},
\end{equation}

Thus, we propose to use the angular prediction as a new predictive form, shown in Eq. \ref{eq:ang}. Intuitively, this form is a non-linear re-weighted form of the original linear prediction. However, this form does not require prior information of the data distribution, is sample-wise, and has its geometric meaning (the angle between vectors). We study the effect of directly replacing the linear predictive form with the angular form in the two stages' testing phase independently. We refer to this method as linear to angular (L2A). 

To be concrete, in stage one we only replace the original linear validation predictive form with the angular form as shown in Eq. \ref{eq:ang}. In stage two, we further omit the linear weight scaling (LWS) module and replace the validation predictive form. The training process of both stages are not modified. This transformation constrains the prediction logits into the range of [0, $\pi$]. Table \ref{tab:cifar} and Table \ref{tab:imagenet_baseline} show the effectiveness of this predictive form. Inspired by the results, we look deeper into this predictive form in the following.

\begin{figure}[tb!]
  \centering
  \includegraphics[width=0.48\linewidth]{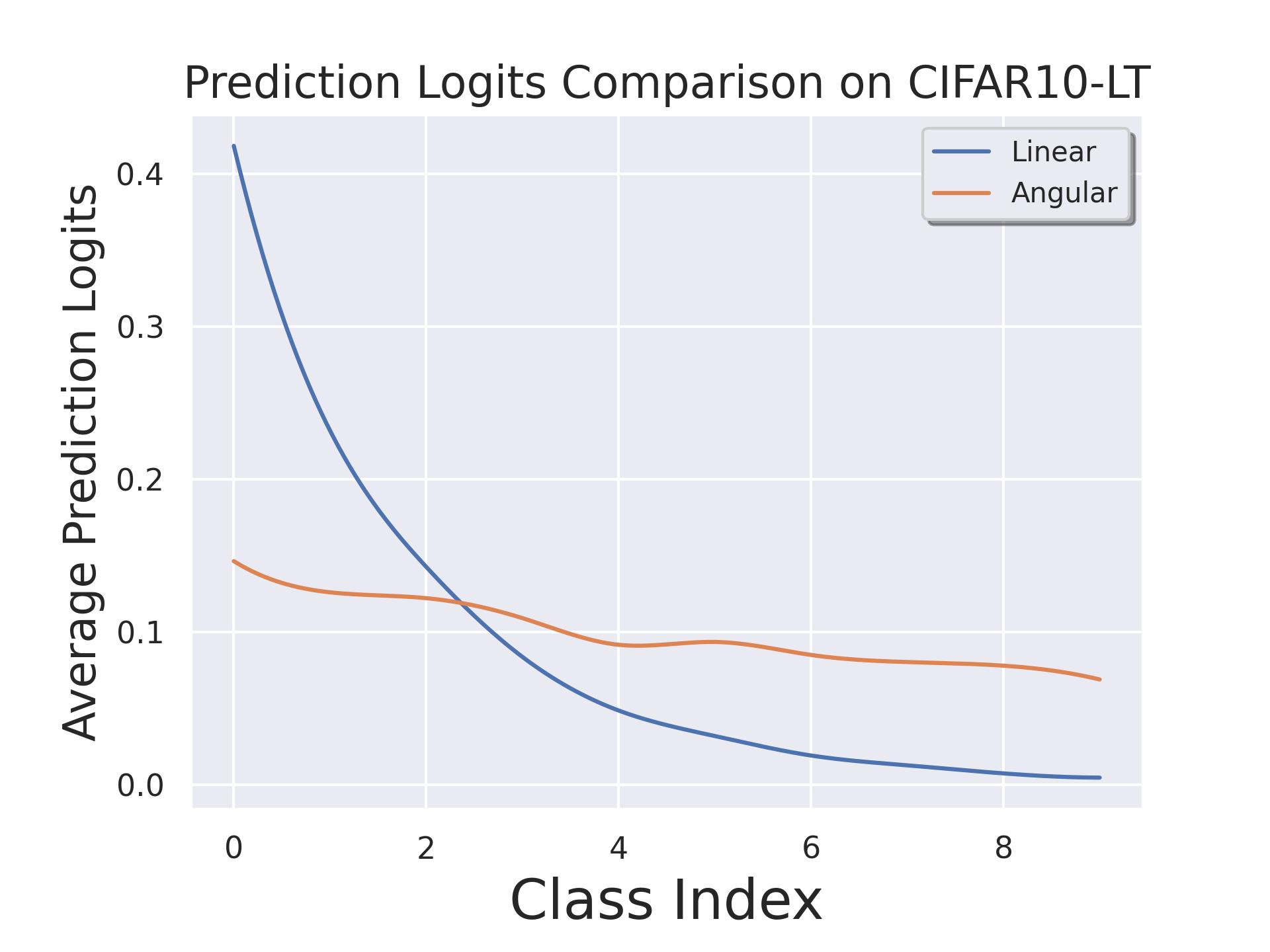}
   \includegraphics[width=0.48\linewidth]{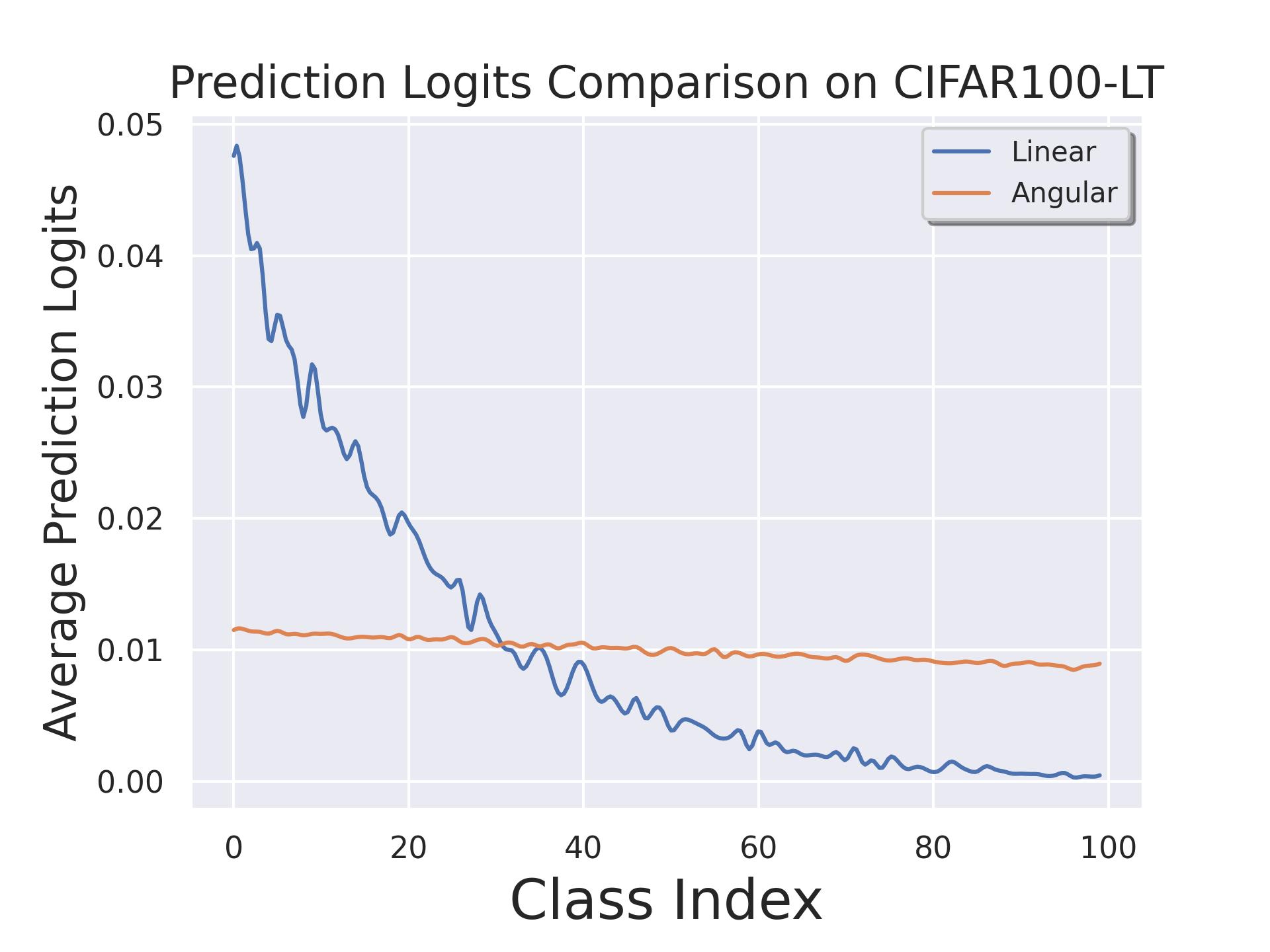}\vspace{-5pt}
   \caption{Comparison of average prediction logits on CIFAR10-LT and CIFAR100-LT. Results are normalized for better visualization. Though both linear and angular prediction logits have a positive correlation with the decreasing sample numbers per class, angular predictions are far smoother.}
   \label{fig:smooth_compare}
   \vspace{-5pt}
\end{figure}

\subsection{Utilizing the Angular Smoothness}
We further discuss the difference between angular predictions and linear predictions in LTR. Fig. \ref{fig:smooth_compare} shows the distribution difference of the two predictive forms. While preserving a similar behavior of being positively correlated with the decreasing sample numbers per class, the angular predictions are much smoother. We refer to this property as \textbf{angular smoothness} and assume the averaged predictions can more effectively reflect the feature space sizes for each class characterized by angles. We show that angular smoothness has its own merits and drawbacks. 

\textbf{Merits:} Specifically, we incorporate the above smoothness into the label-aware smoothing (LAS) module of the second stage of the two-stage learning framework \cite{mislas}. LAS was proposed to resolve the over-confidence in cross-entropy. It uses the prior knowledge of sample numbers per class to construct smoothing factors and reconstruct the one-hot label in the cross-entropy loss:
\begin{equation}
    q_{i} = \left\{
    \begin{aligned}
        &1 - f(N_y), \qquad i=y, \\
        &\frac{f(N_y)}{M-1}, \quad  otherwise,
    \end{aligned}
    \right.
    \label{eq:las_smooth}
\end{equation}
where $q_i$ is the smoothed label for class $i$, $f(N_y)$ is a monotonically decreasing function related to the $y$th class's sample number $N_{y}$. $M$ is the number of classes. LAS assumes the over-confidence is due to the imbalance in sample numbers. By smoothing the labels with respect to different class sample numbers, the tail classes are learned with a higher weights compared to head classes. However, the smoothing factors provide constant weights throughout the learning process, regardless of the actively changing sample batches.

However, LAS's assumption is not flawless. The over-confidence is determined by marginal samples instead of all samples, which accounts for the decision boundary. Angular predictions fall into the limited range of [0, $\pi$], pushing more samples to the decision boundary, and more data points can be seen as marginal samples. Thus we propose active label-aware smoothing (ALAS), which utilizes the angular prediction and takes batch-level information into account. It calculates the batch-wise mean angular prediction results and uses it as a part of the smoothing factor. Denote $\mathcal{P}^{A}(x)$ as the angular probabilistic softmax output of training sample $x$, $\mathcal{P}^{A}_{y}(x)$ is the probability at class index $y$, we calculate the expectation of it over the training samples per batch, and reformulate Eq. \ref{eq:las_smooth} into:
\begin{equation}
    \begin{aligned}
        R^{b}_{y} &= \left(\tau f\left(\frac{1}{B} \sum_{i=1}^{B} \mathcal{P}^{A}_{y}(x_{i})\right) + R^{b-1}_{y}\right) / 2, \\
        q_{i}^{b} &= \left\{
        \begin{aligned}
            &1 - R^{b}_{y}, \quad i=y, \\
            &0, \quad \ otherwise,
        \end{aligned}
        \right.
    \end{aligned}
    \label{eq:alas_smooth}
\end{equation}
where $B$ is the number of samples per batch, $R^{b}_{y}$ is the regularization factor for batch $b$ on class $y$, $R^{0}_{y} = f(N_{y})$. $\tau$ is a hyper-parameter that adjusts the regularization strength of the angular smoothness. We use the same form of $f$ as Eq. \ref{eq:las_smooth} in MiSLAS \cite{mislas}, usually a trigonometric function to normalize the element into the range of [0, 1]. The learnable weight scaling module is also removed when using ALAS.

\textbf{Drawbacks:} However, angular smoothness also introduces some challenges. This over-conservativeness indicates a close-to-uniform distribution space of the predicted logits. Though samples are predicted correctly, the prediction differences between classes are too small, leading to over-pessimistic predictions and uncalibrated results. 

To ease this natural property of angular predictions, we try to increase the variance of the probability distribution. One solution is to increase the entropy of the probabilities, and our new loss for stage one is formulated as:
\begin{equation}
    L(\textbf{q}, \textbf{p}) = - \left(\sum_{i=1}^{M} \textbf{q}_{i}\log \textbf{p}_{i} + \sum_{i=1}^{M} \textbf{p}_{i} \log \textbf{p}_{i}\right)
\label{eq:entloss}
\end{equation}
where $\textbf{q}$ is the ground-truth label, $\textbf{p}$ is the predicted probability, and $M$ is the number of classes. $\textbf{p}$ uses the predictive form of angular prediction $\mathcal{P}^{A}$. We refer to this method as Angular Entropy Minimization (AEM).

\textbf{Composition:} We further combine the two methods into a new training framework. AEM is first used for stage one training with imbalanced data, and ALAS is applied in the second stage learning with balanced data. This framework is referred to as angle-based two-stage learning (ATL), shown in Algorithm \ref{alg:framework}.

\begin{figure}[t]
  \centering
   \includegraphics[width=0.48\linewidth]{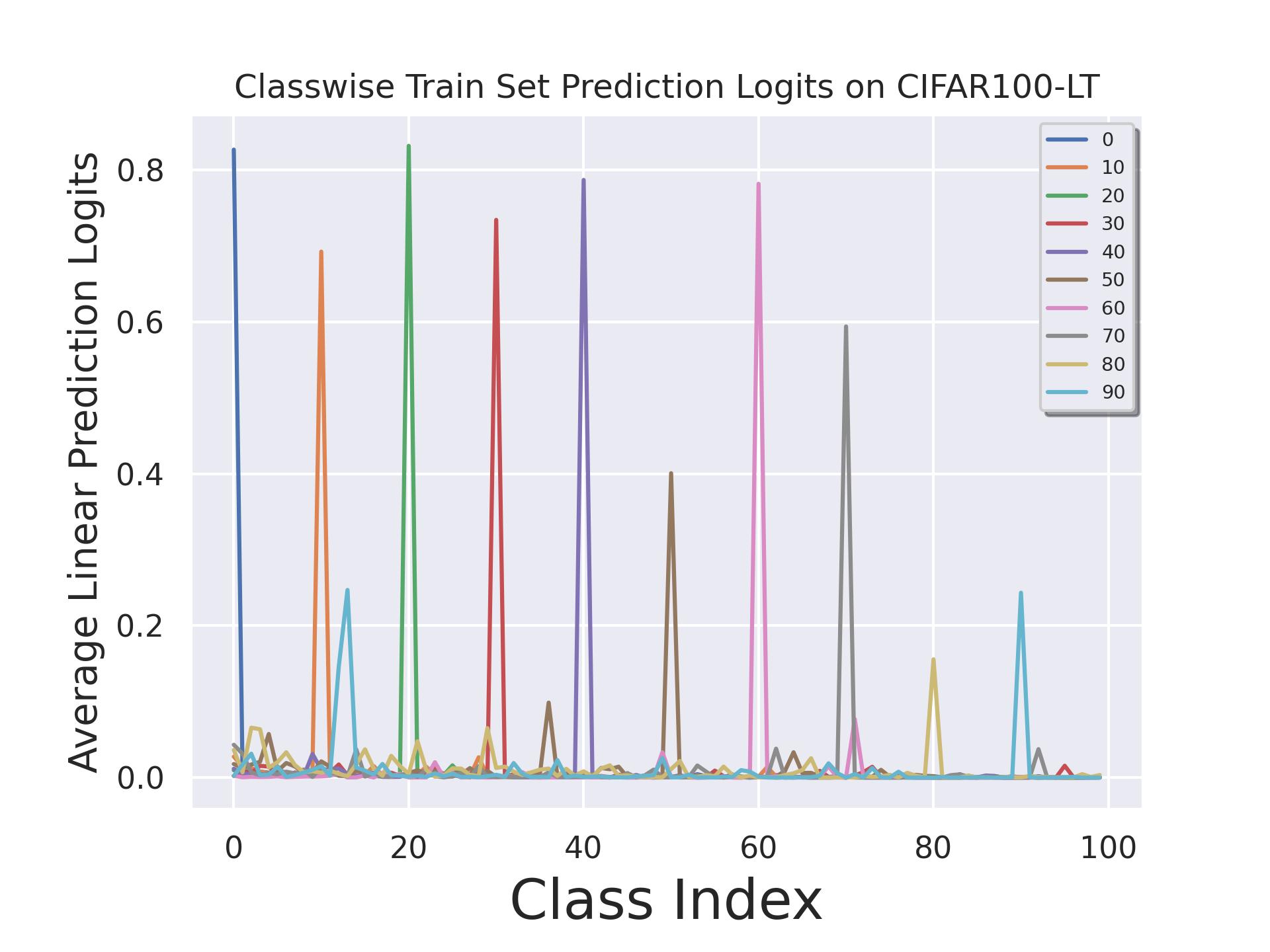}
   \includegraphics[width=0.48\linewidth]{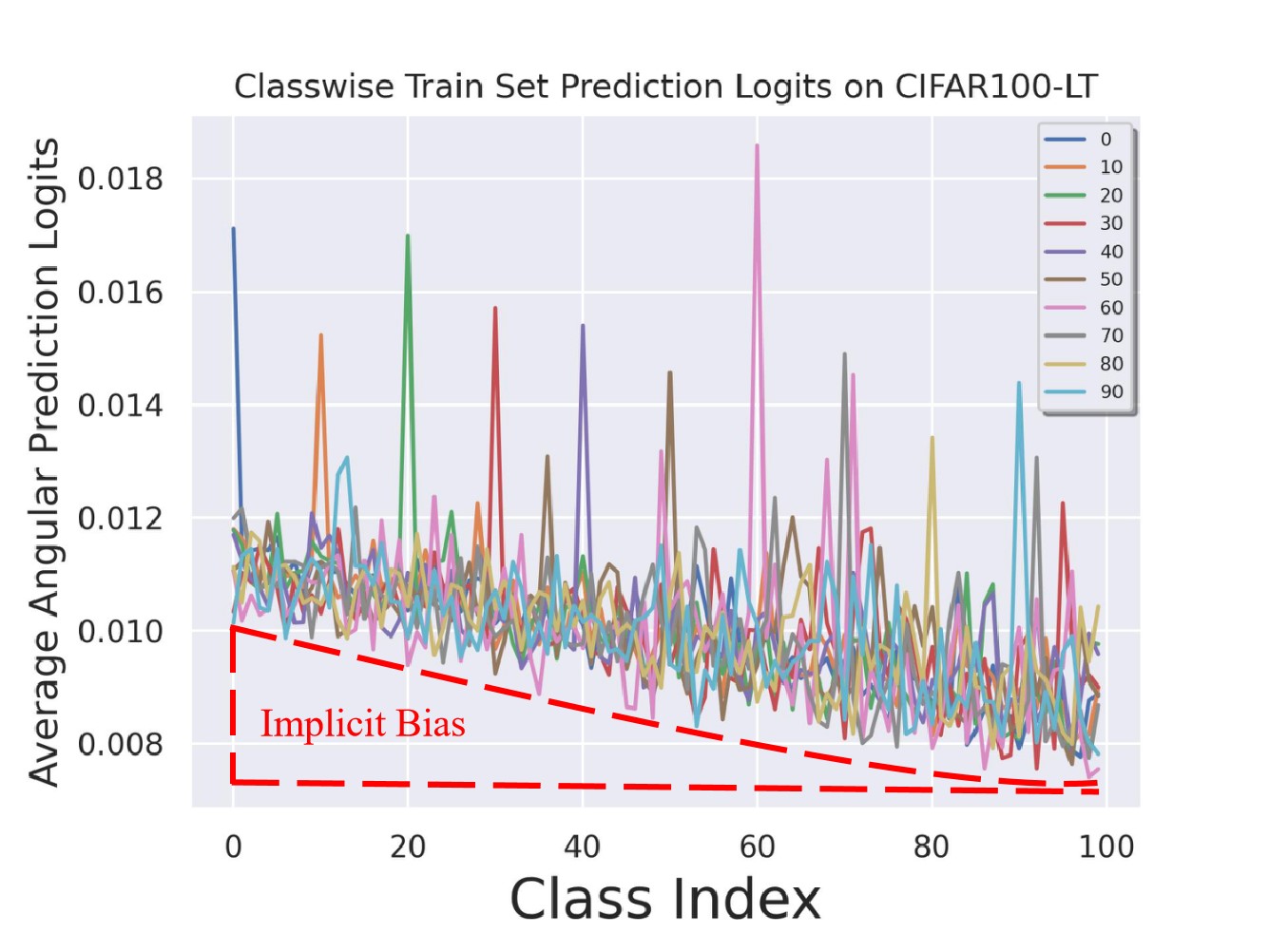}\\
   (a) Linear  \ \ \ \ \ \ \ \ \ \ \ \ \ \ \ \ \ \ \ \ \ \ \ \ \ \ \ \ \ \ \  (b) Angular
   \vspace{-5pt}
   \caption{Class-wise predicted probabilities under different methods on CIFAR100-LT training set. (a) shows that the linear predictions peak at certain classes and press on towards zero at other classes. (b) shows that angular predictions have implicit biases. Results on CIFAR10-LT have a similar distribution.}
   \label{fig:implicit_bias}
\end{figure}

\subsection{Implicit Bias of the Angular Prediction}
In long-tailed recognition, re-weighting methods are not only used in loss modification, but also in changing the prediction results. Learnable weight scaling (LWS) \cite{decoupling} was proposed for stage two learning. It learns a $M$-dimensional vector after the classification layer to directly re-weight the linear prediction probabilities. This re-balancing intend to improve the generalization performance of the model, especially for large scale datasets. The learned scaling parameters monotonically increase with respect to the class index, suppressing the head class probabilities and  expanding the tail class probabilities.

\begin{table*}[tb!]
\caption{Accuracy (\%) comparison on CIFAR datasets. $\beta$ denotes the imbalance ratio. S1 and S2 indicates stage one learning and stage 2 learning, respectively. By simply replacing the linear output with the angular prediction result in the test phase, the prediction accuracy is greatly improved, especially in tail classes. Overall improvements are labeled in bold.}\vspace{-5pt}
\begin{tabular}{ccccccccccccc}
\hline
\multicolumn{1}{c|}{\multirow{3}{*}{Method}} & \multicolumn{12}{c}{CIFAR10-LT}  \\
\multicolumn{1}{c|}{}                        & \multicolumn{4}{c}{$\beta$=100}     & \multicolumn{4}{c}{$\beta$=50}   & \multicolumn{4}{c}{$\beta$=10}  \\
\multicolumn{1}{c|}{}    & Head       & Mid     & Tail     & \underline{Overall} & Head   & Mid    & Tail   & \underline{Overall} & Head      & Mid     & Tail  & \underline{Overall}                   \\ \hline
$\text{MiSLAS}_{S1}$~\cite{mislas}  & 95.27  & 72.96  & 44.77  & 71.19  & 94.94 & 78.43  & 59.83   & 77.81   & 95.58   & 85.48  & 84.28   & 88.15                     \\
$\text{L2A}_{S1}$ (ours) & 93.56  & 77.66  & 62.71  & \textbf{77.94}  & 94.09 & 80.36  & 73.42   & \textbf{82.40}   & 94.29   & 86.30  & 87.78   & \textbf{89.14}  \\
$\text{MiSLAS}_{S2}$~\cite{mislas}  & 91.32  & 80.18  & 72.01  & 81.07   & 91.60   & 82.81  & 81.02  & 84.91  & 92.99   & 87.35  & 90.73  & 90.06    \\
$\text{L2A}_{S2}$ (ours)  & 85.42   & 79.83   & 81.89  & \textbf{82.13}   & 88.94  & 82.23  & 85.94  & \textbf{85.36}   & 91.23  & 87.10   & 92.09   & \textbf{89.84}   \\ \hline
\multicolumn{1}{c|}{\multirow{3}{*}{Method}} & \multicolumn{12}{c}{CIFAR100-LT}                        \\
\multicolumn{1}{c|}{}                        & \multicolumn{4}{c}{$\beta$=100}   & \multicolumn{4}{c}{$\beta$=50}  & \multicolumn{4}{c}{$\beta$=10}   \\
\multicolumn{1}{c|}{}  & Head  & Mid  & Tail & \underline{Overall} & Head   & Mid     & Tail    & \underline{Overall} & Head     & Mid     & Tail  & \underline{Overall}   \\ \hline
$\text{MiSLAS}_{S1}$~\cite{mislas}   & 68.00   & 36.51  & 5.55    & 38.87    & 69.35  & 44.04   & 12.09    & 43.89  & 71.88   & 58.61    & 38.99  & 57.70  \\
$\text{L2A}_{S1}$ & 67.06 & 38.14 & 8.07 & \textbf{39.83}  & 68.33 & 43.97 & 15.90 & \textbf{44.60} & 70.86 & 59.66 & 40.59 & \textbf{58.16} \\
$\text{MiSLAS}_{S2}$~\cite{mislas}  & 62.39 & 47.50 & 24.00 & 46.05 & 62.20 & 53.37 & 35.10 & 51.25 & 66.11 & 62.26 & 55.15 & 61.58 \\
$\text{L2A}_{S2}$ & 59.56 & 44.89 & 27.79 & \textbf{45.21} & 60.25 & 48.06 & 37.38 & \textbf{49.35} & 65.81 & 61.60 & 55.72 & \textbf{61.41} \\ \hline
\end{tabular}
\label{tab:cifar}
\end{table*}

Different from the linear prediction, angular prediction yields a different characteristic. Fig.~\ref{fig:implicit_bias} shows the mean predicted probabilities for each class on CIFAR100-LT using different methods. Fig.~\ref{fig:implicit_bias}(a) shows that the linear predictions are close to zero probabilities on most classes, and peaks at few classes. However, for angular predictions, Fig.~\ref{fig:implicit_bias}(b) indicates that there are prediction biases existent. Moreover, the bias is monotonically increasing with respect to the class sample numbers. We refer to this phenomenon as the implicit bias for angular predictions. Note that in the plots, we sample 10 out of 100 classes for clearer visualization, and does not effect our observations. Plots on CIFAR10-LT and CIFAR100-LT with all classes are shown in the supplementary materials.

We use it as an indicator of the prediction bias on test data, and calibrate it through logits re-weighting. We refer to this method as Angular Bias-directed Smoothing (ABS). The calibration on test data $x$ is done by:
\begin{equation}
\begin{aligned}
     &\mathcal{P}_{te}(x) = \gamma \mathcal{P}_{te}(x), \\ 
    \text{where} \ \ &\gamma = 1 - s * \mathcal{F}(\mathcal{P}_{tr}).
\end{aligned}
\label{eq:bias_smooth}
\end{equation}
$\mathcal{P}_{te}(x)$ is the softmax output of the testing data $x$ and $\mathcal{P}_{tr}$ is the softmax output of all training data. $\gamma$ is the re-weighting factor of $M$-dimension. $\mathcal{F}$ is an monotonically decreasing function that uses the train data predictions. $s$ is a hyper-parameter that adjusts the magnitude of $\mathcal{F}$, falling into the range of $[0, 1]$. To be concrete, $\mathcal{F}$ first calculates the mean prediction logits $\mathbb{E}[\mathcal{P}_{tr}^{c}]$ of the training data for each class $c$, then calculates the minimum confidence for each class: $\mathcal{B}_{tr} = \text{min}_{i \in [1, C]}\mathbb{E}[\mathcal{P}_{tr}^{i}]$, obtaining a $C$-dimensional vector, where $C$ is the number of classes. Finally, we formulate the normalized prediction logits to characterize the re-weighting distribution. Two distribution forms are used:
\begin{itemize}
    \item Sine:
        \begin{equation}
            \mathcal{F}(\mathcal{P}_{tr}) = \sin \left(\frac{\pi * (\mathbb{E}[\mathcal{B}_{tr}] - \text{min}(\mathbb{E}[\mathcal{B}_{tr}]))}{2 * (\text{max}(\mathbb{E}[\mathcal{B}_{tr}]) - \text{min}(\mathbb{E}[\mathcal{B}_{tr}]))} \right),
            \label{eq:dis_form_concave}
        \end{equation}
    \item Linear:
    \begin{equation}
            \mathcal{F}(\mathcal{P}_{tr}) = \frac{\mathbb{E}[\mathcal{B}_{tr}] - \text{min}(\mathbb{E}[\mathcal{B}_{tr}])}{\text{max}(\mathbb{E}[\mathcal{B}_{tr}]) - \text{min}(\mathbb{E}[\mathcal{B}_{tr}])}.
            \label{eq:dis_form_linear}
    \end{equation}
\end{itemize}

\begin{figure*}[t]
  \centering
   \includegraphics[width=0.33\linewidth]{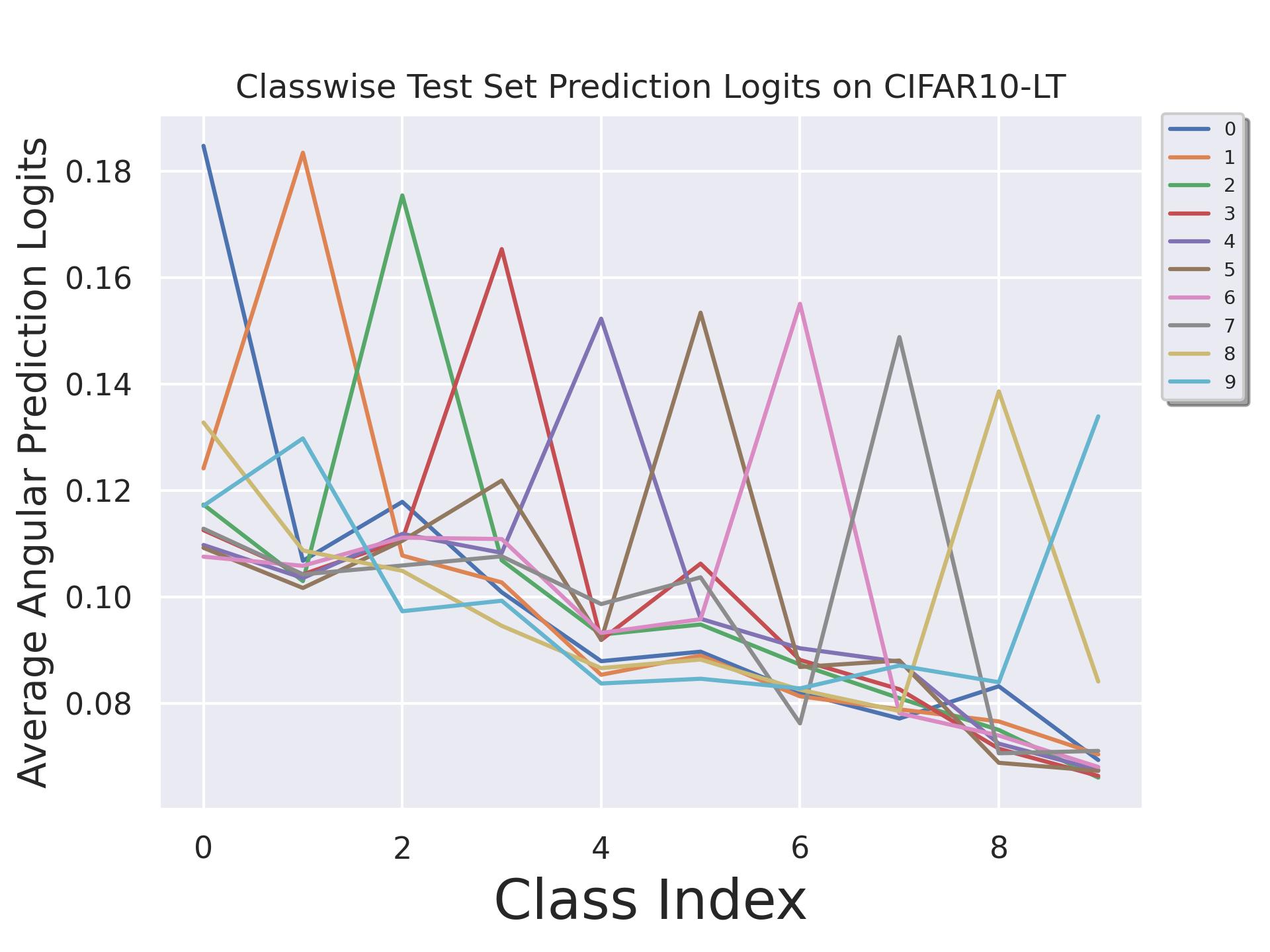}
   \includegraphics[width=0.33\linewidth]{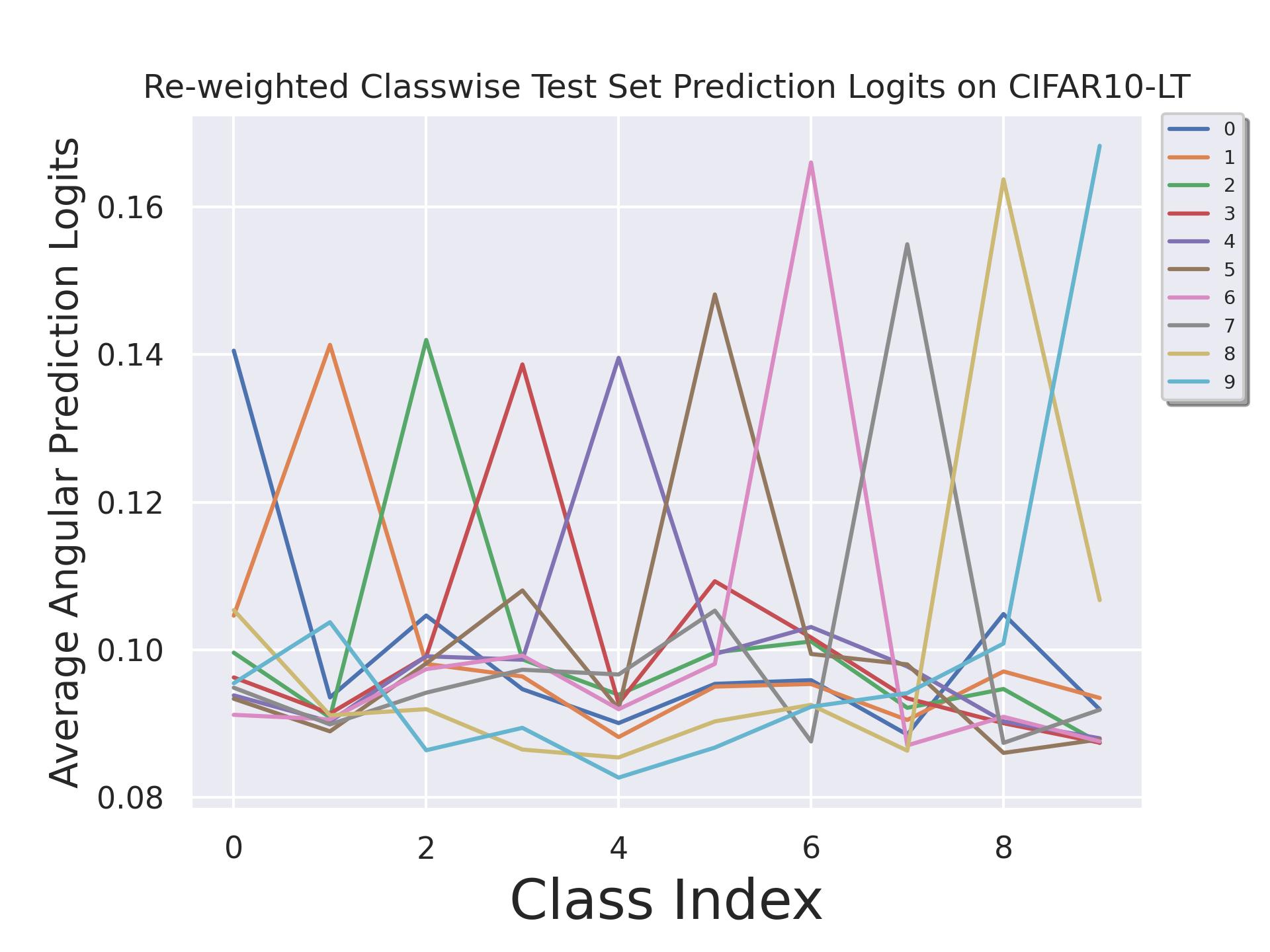} 
   \includegraphics[width=0.33\linewidth] {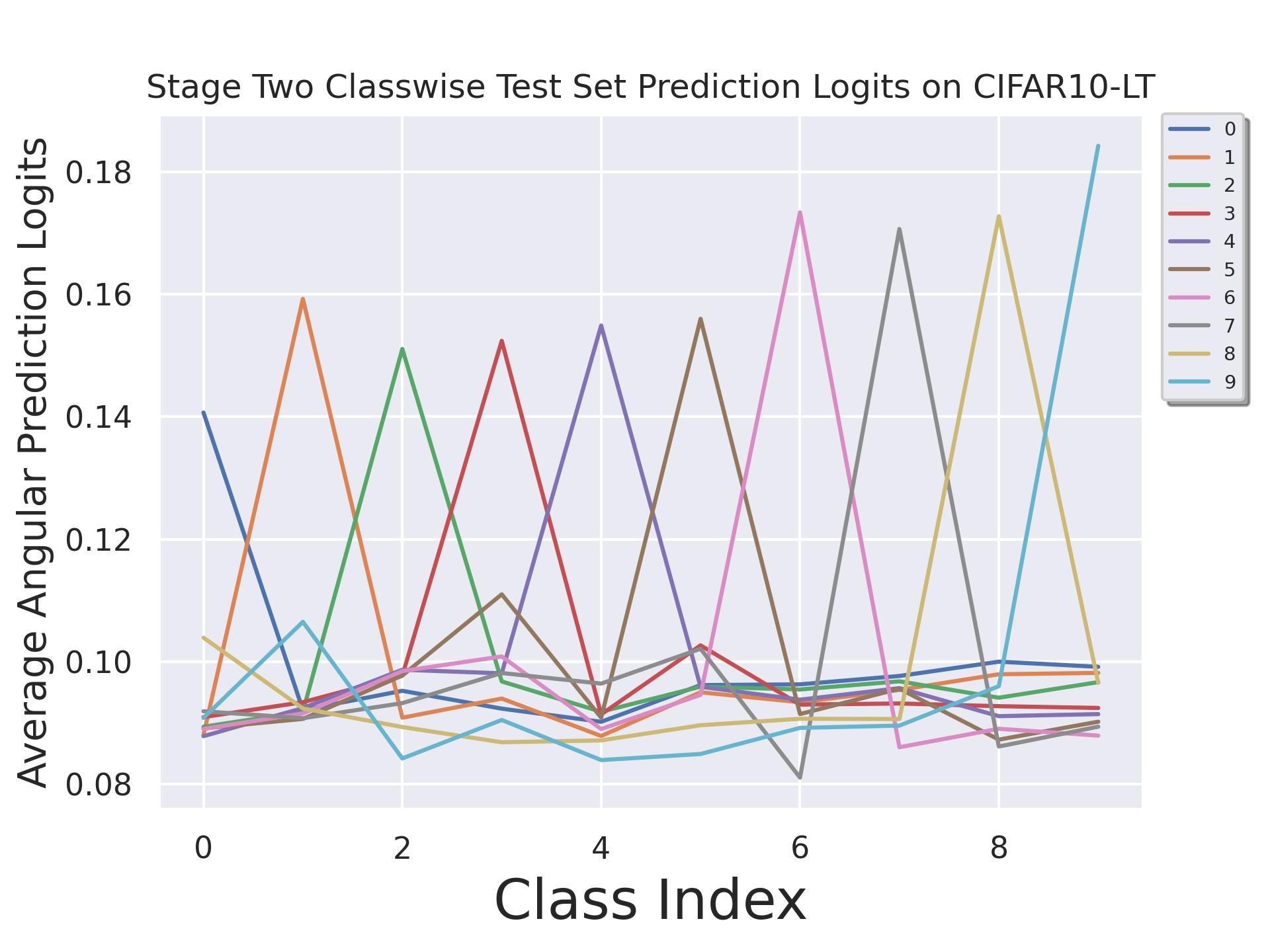} \\
   (a) Existent implicit bias \qquad \qquad \qquad \qquad (b) Elimination by ABS \qquad  \qquad \quad \ \  (c) Elimination by MiSLAS
   \vspace{-5pt}
   \caption{Class-wise predicted probability comparison on CIFAR10-LT test set. (a) presents the results before re-weighting; (b) shows the results after applying our re-weighting method on the predicted logits; (c) uses the second stage training process and is seen as the ground-truth, representing a well-calibrated prediction. We show that on CIFAR10-LT, probability re-weighting using angular prediction results can decrease the implicit bias and even approximate the second stage learning.}
   \label{fig:exp_reweighted}
\end{figure*}

\begin{algorithm}[t]
\caption{Angle-based Two-stage Learning (ATL)}
\label{alg:framework}
\begin{small}
\begin{algorithmic}[1]
\STATE {\bfseries Input}: Long-tailed data $\mathcal{D}_{l}$, balanced data $\mathcal{D}_{b}$, feature extractor $\mathcal{F}_{e}$, classifier $\mathcal{F}_{c}$
\STATE {\bfseries Stage one}: Calculate angular prediction $\mathcal{P}^{A}$ in each epoch using $\mathcal{F}_{e}$'s feature output and $\mathcal{F}_{c}$'s weights via Eq. \ref{eq:ang}
\STATE  \qquad \qquad \ \ Train $\mathcal{F}_{e}$ and $\mathcal{F}_{c}$ on $\mathcal{D}_{l}$ using Eq. \ref{eq:entloss} with $\mathcal{P}^{A}$
\STATE {\bfseries Stage two}: Freeze $\mathcal{F}_{e}$, finetune $\mathcal{F}_{c}$ on $\mathcal{D}_{b}$ as follows:
\STATE  \qquad For each epoch $e$:
\STATE  \qquad \qquad Calculate the regularization factor $\mathcal{R}_{y}^{e}$ in Eq. \ref{eq:alas_smooth} with the $\mathcal{R}_{y}^{e-1}$ in last epoch and calculated $\mathcal{P}^{A}$
\STATE  \qquad \qquad Take the negative log-likelihood as the loss using 
\qquad  \qquad \quad smoothed label ($1 - \mathcal{R}_{y}^{e}$)
\STATE {\bfseries Output}: Trained feature extractor $\mathcal{F}_{e}$ and classifier $\mathcal{F}_{c}$
\end{algorithmic}
\end{small}
\end{algorithm}

\section{Experiments and Analysis}
\subsection{Experimental Settings}
We provide empirical results on different long-tailed recognition datasets, including CIFAR10-LT, CIFAR100-LT and ImageNet-LT. On CIFAR10-LT and CIFAR100-LT, we sample images from the original balanced versions \cite{cifar100} \cite{imagenet} using an exponential distribution:
\begin{equation}
    n_{c} = \frac{N}{M} * \beta^{\frac{c}{M-1}}
\end{equation}
where $n_{c}$ denotes the sample number for class $c$, $N$ is total number of images in the original dataset, $M$ is the total number of images and $\beta$ is the imbalance ratio. The imbalance ratio here is defined as the quotient of the maximum class sample number over the minimum class sample number. We use different imbalance ratios $\beta$ of 100, 50 and 10. ImageNet-LT contains 1,000 classes with sample numbers ranging from 5 to 1,280. 

For fair comparisons with different baselines, we align our settings with MiSLAS. On CIFAR10-LT and CIFAR100-LT, we use the CIFAR style ResNet32. On ImageNet-LT, we use ResNet50 \cite{resnet}. The backbones are trained from scratch without loading pretrained parameters. Accuracy is chosen as our evaluation protocol. Regarding the large number of classes, we further divide the categories into head classes, middle classes and tail classes. On CIFAR10-LT, the class indexes for head, middle and tail classes are in the range of $[0, 3)$, $[3, 7)$ and $[7, 10)$, respectively. On CIFAR100-LT, the ranges are $[0, 36)$, $[36, 71)$ and $[71, 100)$ respectively. On ImageNet-LT, the ranges are $[0, 390)$, $[390, 835)$, $[835, 1000)$. Due to the different range settings adopted by different works, we only compare the head, middle and tail class performances with MiSLAS. 

Our implementations are based on the PyTorch toolbox \cite{pytorch}. We use a maximum of 4 Tesla V100 GPUs. ResNet32 models are trained for 300 epochs with batch size of 128, ResNet50 is trained for 180 epochs with batch size of 256. We use the SGD~\cite{sgd} optimizer with learning rate of 0.1, momentum of 0.9 and weight decay rate of 0.0005. 

\begin{table}[tb!]
\centering\caption{Performance (\%) comparison on CIFAR10-LT ($\beta=100$) and CIFAR100-LT ($\beta$=100) with different baselines. $\text{MiSLAS}_{S1}$ and $\text{MiSLAS}_{S2}$ indicates the one and two stage training respectively, and the results are re-implemented by us because our methods are based on it. Best performance viewed in bold.}\vspace{-5pt}
\begin{tabular}{rcc}
\hline
Accuracy (top-1)  & CIFAR10-LT & CIFAR100-LT  \\ 
\hline \hline
\multicolumn{3}{c}{Baselines}   \\ \hline 
Focal Loss \cite{focalloss}  & 70.3   & 38.4  \\ \hline
L2RW  \cite{L2RW}      & 74.1   & 40.2  \\ \hline
CB Loss \cite{cbloss}    & 74.5   & 39.6  \\ \hline
RCBM-CE \cite{RCBM-CE}    & 76.4   & 43.3  \\ \hline
BBN  \cite{bbn}       & 79.8   & 42.5  \\ \hline
TSC  \cite{TSC}       & 79.7   & 43.8   \\ \hline
TDE \cite{tde}        & 80.6   & 44.1   \\ \hline
$\text{MiSLAS}_{S1}$~\cite{mislas}  & 71.19 & 38.87 \\ \hline
$\text{MiSLAS}_{S2}$~\cite{mislas}  & 81.07 & 46.05 \\ \hline \hline
\multicolumn{3}{c}{Ours}        \\ \hline
$\text{L2A}_{S1}$  & 77.94  & 39.83  \\ 
$\text{L2A}_{S2}$   & 82.13  & 45.21  \\ \hline
ABS    &  82.01 &   43.88  \\ \hline
$\text{ATL}_{AEM}$    &  78.23  &  40.77   \\ 
$\text{ATL}_{ALAS}$   &  82.35  &  46.54  \\ 
$\text{ATL}_{all}$ & \textbf{82.60} & \textbf{47.11}  \\ \hline
\end{tabular}
\label{tab:cifar_baseline}
\end{table}

\subsection{Empirical Analysis on Angular Information}
We study the effects of different usages of angular information, including (1) Replacement of the linear prediction using angular prediction; (2) Smoothness of the  prediction; (3) Implicit bias elimination of the prediction. For the baselines included, we select the most relevant and fair works with the same backbones.

\textbf{Replacing the linear form:} Table \ref{tab:cifar} and Table \ref{tab:imagenet_baseline} show our results. We perform the experiments for three times with different random seeds and calculate the mean of the last epoch's performance. From the two tables, it is observed that the angular prediction is superior to linear prediction results in long-tailed recognition, especially in tail class performance. We also observe that the gain in performance of stage one is significant, and is larger than that in stage two. L2A is able to exclude the linear weight scaling module in stage two, saving the number of parameters needed to be learned. 

\textbf{Exploiting the prediction smoothness:} Table \ref{tab:cifar} and Table \ref{tab:imagenet_baseline} show results for using angle-based two-stage learning (ATL). $\text{ATL}_{AEM}$ means using angular entropy minimization individually and $\text{ATL}_{ALAS}$ means using only active label-aware smoothing. Note that for ALAS, we freeze the feature extraction layers and only finetune the classification layer, so as to be consistent with the LAS learning criterion. From the tables, we see that both methods are able to improve the performance, indicating that introducing entropy is beneficial and ALAS is better than the original smoothing factor. We also show that using the whole framework ($\text{ATL}_{all}$) further benefits long-tailed recognition.

\textbf{Elimination of the implicit bias:} The mean class-wise prediction results for CIFAR10-LT test set are shown in Fig. \ref{fig:exp_reweighted}. The comparison plots for CIFAR100-LT (sampling 10 out of 100 classes) are shown in the supplementary materials. For CIFAR10-LT, we use the concave form of $\mathcal{F}$ with $s=0.25$, and on CIFAR100-LT, the choice is the linear form with $s=0.1$. Further ablation studies for this hyper-parameter is provided in Section \ref{sec:ab_study}.

Fig.~\ref{fig:exp_reweighted}(a) plots the mean angular prediction logits distribution after the first stage training on the validation set, showing a similar trend with the training set. The results for further using stage two balanced data classifier re-training is shown in Fig.~\ref{fig:exp_reweighted}(c), whose distributions are more calibrated. Fig.~\ref{fig:exp_reweighted}(b) visualizes the logits distribution after our re-weighting strategy, which eliminates the implicit bias and resembles the distribution in (c). Table \ref{tab:cifar_baseline} and Table \ref{tab:imagenet_baseline} shows that re-weighting is able to improve the baseline. These experimental results show that our re-weighting method that utilizes the angular prediction results can ease the implicit bias, and even approximate the performance of two stage training without re-sampling the long-tailed data.

\begin{table}[tb!]
\centering
\caption{Performance (\%) comparison on ImageNet-LT. Best viewed in bold. $\text{MiSLAS}_{S1}$ and $\text{MiSLAS}_{S2}$ are re-implemented.}\vspace{-5pt}
\begin{tabular}{rc}
\hline
Accuracy (top-1) & ImageNet-LT \\ \hline \hline
\multicolumn{2}{c}{Baselines}        \\ \hline
MetaSAug-CE \cite{metasaug} &  47.30  \\
LWS \cite{decoupling}       &  49.90  \\
$\text{MiSLAS}_{S1}$~\cite{mislas}     &  44.34  \\
$\text{MiSLAS}_{S2}$~\cite{mislas}     &  51.16   \\ \hline \hline
\multicolumn{2}{c}{Ours}        \\ \hline
$\text{L2A}_{S1}$    &   49.53  \\ 
$\text{L2A}_{S2}$    &   50.66  \\ \hline
ABS       &   50.00  \\  \hline
$\text{ATL}_{AEM}$       &   49.35  \\ 
$\text{ATL}_{ALAS}$     &   51.28  \\ 
$\text{ATL}_{all}$ &  \textbf{51.37}   \\ \hline
\end{tabular}
\label{tab:imagenet_baseline}
\end{table}

\subsection{Extending the Applications}
Apart from aiding the two-stage learning in long-tailed recognition, we also explore the possible aspects that angular predictions are useful. We first discuss the change of backbones from CNNs to vision transformers. Then we explore the exciting area of data pruning for long-tailed data.

\begin{table}[tb!]
\centering
\caption{Performance (\%) on ImageNet-LT using ViT backbones.}\vspace{-5pt}
\begin{tabular}{c|c|llllllll}
\hline
\multicolumn{2}{c}{Accuracy} & Head & Mid & Tail & \underline{Overall} \\ \hline \hline
\multirow{2}{*}{ViT-S} & Linear & 82.71 & 67.98 & 41.39 & 69.24   \\
   & Angular & 82.12 & 66.29$\downarrow$ & 43.29 $\uparrow$ & 68.66   \\ \hline 
\multirow{2}{*}{ViT-B} & Linear & 84.99 & 70.32 & 44.93 & 71.85  \\ 
    & Angular & 84.32 &  63.02$\downarrow$ &  47.82 $\uparrow$ & 71.05  \\ \hline
\end{tabular}
\label{tab:vit}
\end{table}

\textbf{Vision Transformer as Backbone:} Vision transformers (ViTs) \cite{vit} have proved its superiority in both language and image tasks, benefited from its attention mechanism and the large amount of data pre-trained on. However, angular information is rarely discussed using ViTs, works such as \cite{decoupledNet} and \cite{avh} only probe into the CNN setting. In the supplementary materials, we show that the weight imbalance does not exist in ViT's head classifier. We look into this area to see how angular prediction works. 

Table \ref{tab:vit} shows our performance on ImageNet-LT using ViT-S and ViT-B. We train the models under the Deit \cite{deit} fine-tuning setting and only replace the original ImageNet with the long-tailed version. When using L2A, we can observe an competitive performance with the linear output, showing a tradeoff between the middle and tail class accuracy.

\begin{figure}[t!]
  \centering
   \includegraphics[width=0.48\linewidth]{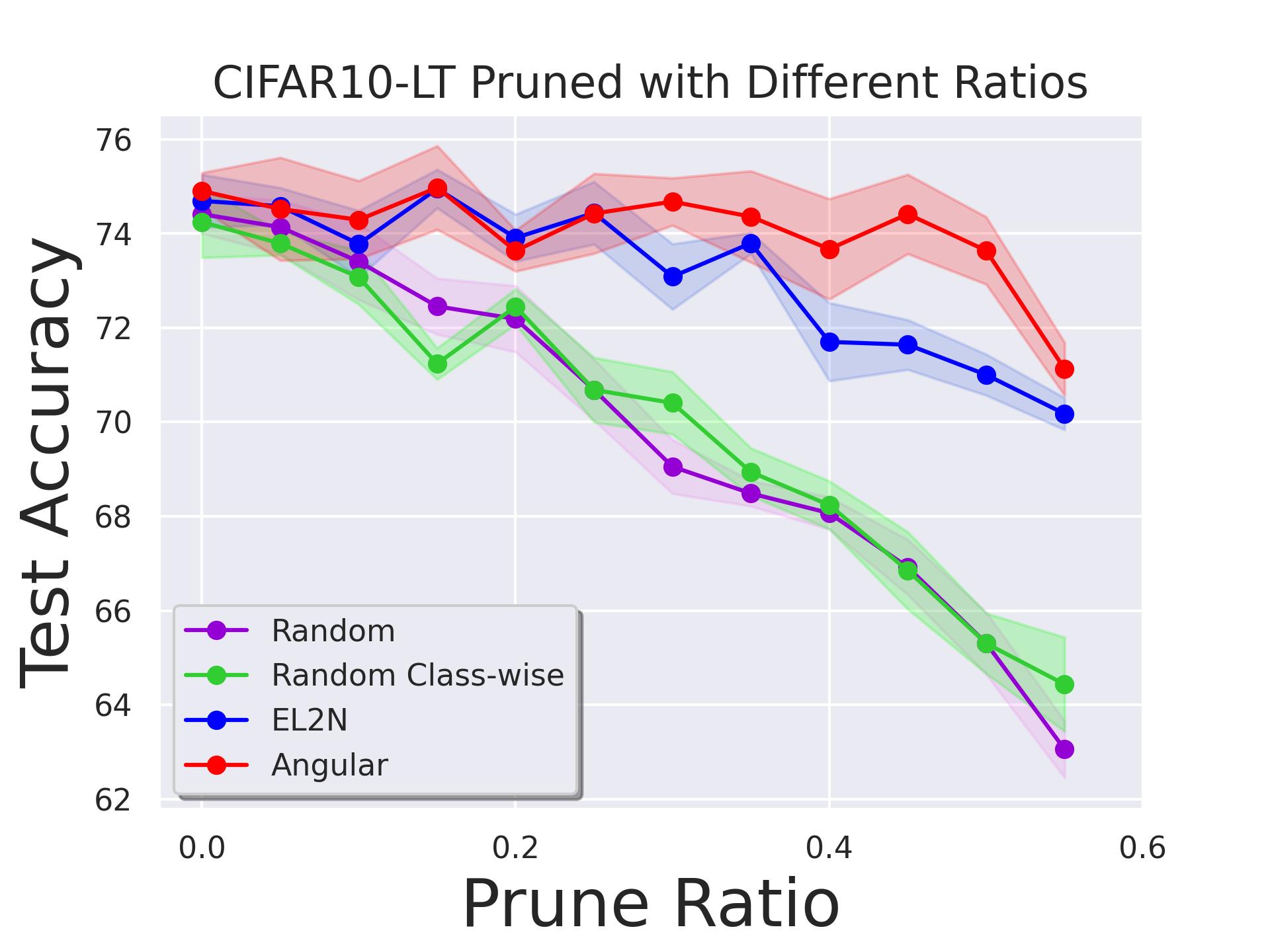}
   \includegraphics[width=0.48\linewidth]{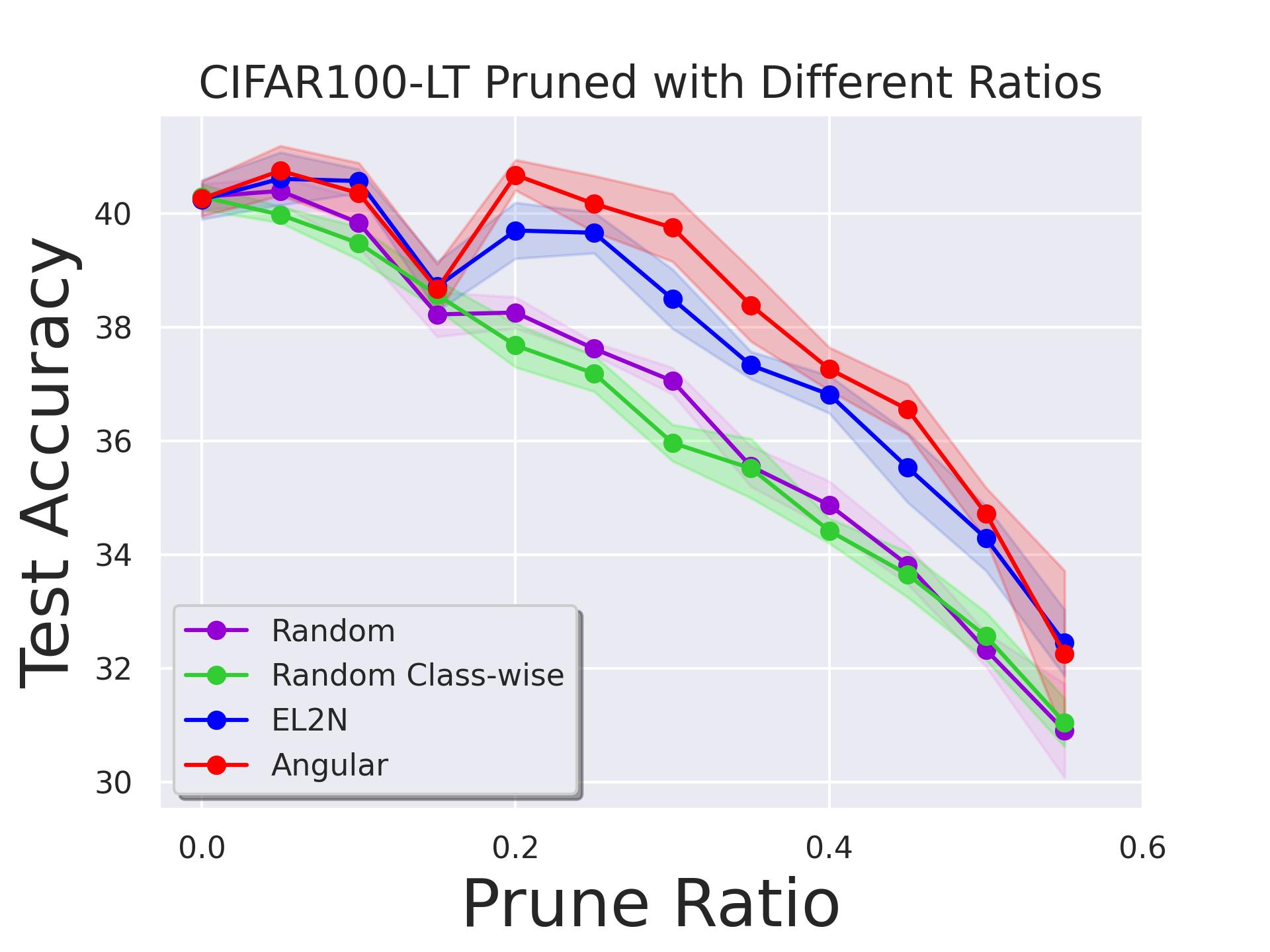}\vspace{-5pt}
   \caption{Data pruning results on CIFAR10-LT and CIFAR100-LT. AVH is able to achieve the best results, being able to conserve up to $45\%$ training data on CIFAR10-LT and $25\%$ data on CIFAR100-LT without only little loss in performance.}
   \label{fig:exp_data_prune}
\end{figure}
\begin{table*}[tb!]
\centering
\caption{Performance (\%) on CIFAR10-LT and CIFAR100-LT under different $s$ in Eq. \ref{eq:bias_smooth}, which accounts for the trade-off between the train and validation set. Two forms of $\mathcal{F}(\mathcal{P}_{tr})$ (Eq. \ref{eq:dis_form_concave}, Eq. \ref{eq:dis_form_linear}) are also compared.}
\vspace{-5pt}
\begin{tabular}{c|ccccccccc}
\hline
$s$           & 0 & 0.04 & 0.08 & 0.12 & 0.16 & 0.20 & 0.24 & 0.28 & 0.32 \\ \hline \hline
CIFAR10-LT (sine) & 76.78 & 77.90 & 79.18 & 80.39 & 81.56 & 81.67 & 82.01 & 81.43 & 80.41 \\ \hline 
CIFAR100-LT (linear) & 39.83 & 41.14 & 42.46 & 42.97 & 43.65 & 43.88 & 42.92 & 41.22 & 38.59 \\ \hline
\end{tabular}
\label{tab:ablation_s}
\end{table*}

\begin{table}[tb!]
\centering
\caption{Performance (\%) on CIFAR100-LT under different $\tau$ in Eq. \ref{eq:alas_smooth} (controlling the trade-off between the head and tail classes performance) and different choices of $f(\cdot)$: linear and concave.}\vspace{-5pt}
\begin{tabular}{c|ccccccccc}
\hline
$\tau$ & 0.25 & 0.50 & 0.75 & 1.00 & 1.25 \\ \hline \hline
Linear  & 46.31 & 46.52 & 46.50 & 46.18 & 45.85  \\ \hline 
Concave  & 46.01 & 46.29 & 46.54 & 46.38 & 45.93  \\ \hline
\end{tabular}
\label{tab:ablation_tau}
\end{table}
\textbf{Long-tailed Data Pruning:} Data pruning removes some samples from the training set to compress the dataset size, for the goal of reducing the model training cost and even improve the generalization performance. The key idea for the pruned samples is that they are easier than others and are not necessary for the learning process, or they are outliers/noises that damage the training procedure. We extend this setting to long-tailed data, as we assume large scale datasets are more easily to have imbalanced distributions, and long-tailed data pruning would be a more general and challenging topic.   

A common strategy for data pruning is to assign each training sample with a score calculated by a certain metric, then the top/bottom percentage of samples are pruned from the training set. A popular and effective pruning metric is EL2N \cite{el2n}, which trains the same model with multiple random seeds for few epochs, and calculates the L2 norm of the error vector. The EL2N score can be calculated by:
\begin{equation}
    \mathcal{S}_{EL2N}(x) = \mathbb{E}[||\mathcal{P}(x) - \mathcal{T}(x)||_{2}]
    \label{eq:el2n}
\end{equation}
where $\mathcal{S}_{EL2N}$ is the EL2N score, $\mathcal{P}(x)$ is the linear prediction output for sample $x$, $\mathcal{T}(x)$ is the corresponding one-hot label. We instead use the angular visual hardness score $\mathcal{S}_{AVH}$ proposed by \cite{avh}, which is in the form of:
\begin{equation}
    \begin{aligned}
        &\mathcal{A}_{i}(x) = \arccos\left(\frac{\phi(x) \cdot W_{i}}{||\phi(x)||||W_{i}||}\right) \\
        &\mathcal{S}_{AVH}(x) = \mathbb{E}\left[\frac{\mathcal{A}_{y}}{\sum_{i}\mathcal{A}_{i}}\right]
    \end{aligned}
    \label{eq:avh}
\end{equation}

We compare different methods on CIFAR10-LT and CIFAR100-LT. The baselines include: (1) Random sampling, which randomly prune a certain proportion of data from the whole training set; (2) Class-wise random sampling, which randomly prune the same proportion of data from each class; (3) EL2N; (4) AVH. For EL2N and AVH, we train ten ResNet32 models initialized with different random seeds for ten epochs, freeze the model parameters and calculate the corresponding scores. The experiments are repeated with 5 different random seeds and we calculate the mean and variance. The results are shown in Fig.~\ref{fig:exp_data_prune}.

From the results, we show that AVH is superior compared with other baselines. AVH can have only little test accuracy loss even with 45\% data pruned on CIFAR10-LT, and 25\% data pruned on CIFAR100-LT. The performance superiority can be regarded as the advantage introduced by angular information. EL2N's prediction logits are effected by the imbalance in classifier weights, leading to errors in determining sample hardness, thus incorrectly pruning the necessary samples (hard ones) from the training set. AVH, on the other hand, removes the negative effect of the classifier imbalance. Also, as discussed by \cite{avh}, AVH is able to resemble the human perception and visual hardness, leading to better identification of the samples' difficulties. 


\subsection{Ablation Study}
\label{sec:ab_study}
\vspace{-5pt}
We discuss the effects of hyper-parameter, including the selection for $\tau$ in Eq. \ref{eq:alas_smooth} and $s$ in Eq. \ref{eq:bias_smooth}.

\textbf{Choices for $\tau$:} Though both used as smoothing factors, a natural gap exists between prediction logits and number of samples per class. Thus we use $\tau$ to balance the two elements. $f(\cdot)$ has two forms, which are the same as in Eq. \ref{eq:dis_form_concave} and Eq. \ref{eq:dis_form_linear}, with the concave form as $f(x)=sin(\pi * (x-\text{min}(x)) / 2*(\text{max}(x)-\text{min}(x)))$, and the linear form as $f(x)=(x-\text{min}(x)) / (\text{max}(x)-\text{min}(x))$. From Table \ref{tab:ablation_tau}, we observe that the linear form has a better performance when $\tau$ is smaller, and the concave form is better when $\tau$ is larger. Moreover, the accuracy-$\tau$ curve has a convex distribution. Particularly, the distributions in the middle and tail classes are also convex, while in the head classes it is concave.

\textbf{Choices for $s$:} $s$ adjusts the magnitude of re-weighting to ease the implicit bias. Table \ref{tab:ablation_s} shows the ablation study on two datasets. From the table, we can see that the max accuracy performance is achieved at $0.25$ and $0.1$, due to the trade-off in accuracy between the head and tail classes.

\section{Conclusion}
We have proposed a feature-weight angular prediction based classifier output reformulation approach to avoid the weight imbalance issue encountered in LTR. Extensive experimental results on benchmarks show our method outperforms other SOTA methods without pretraining, and performs closely to those contrastive methods with pretraining. Our approach can also serve as plugin to be adapted to vision transformers and data pruning, to improve their performance on LTR notably.


{\small
\bibliographystyle{ieee_fullname}
\bibliography{egbib}
}


\newpage
\appendix
\section{Additional Visualization}
In this section of the supplement, we provide additional visualization of the plots in the main paper. 

\subsection{Angular smoothness}
\begin{figure}[h]
  \centering
  \includegraphics[width=0.48\linewidth]{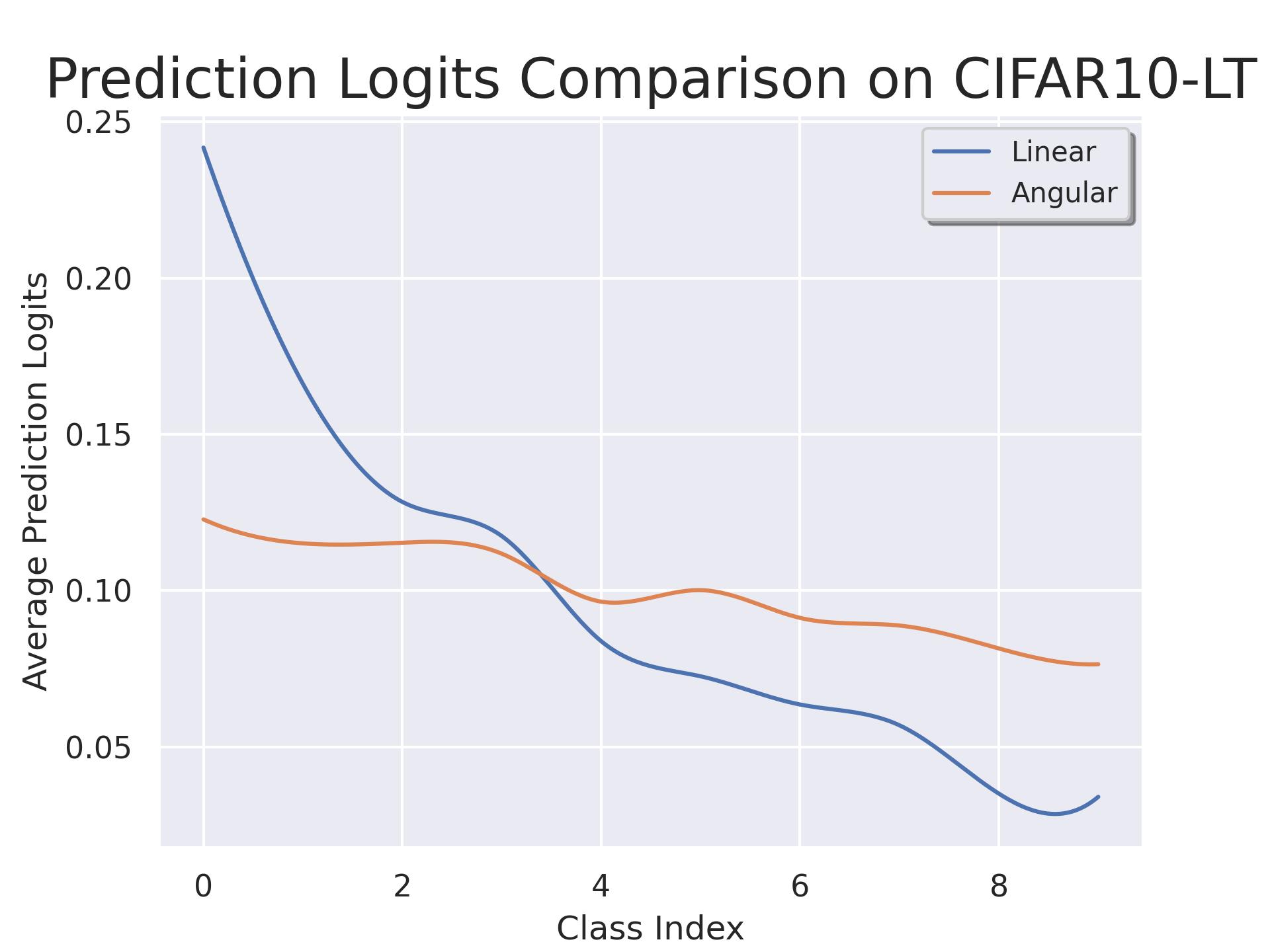}
   \includegraphics[width=0.48\linewidth]{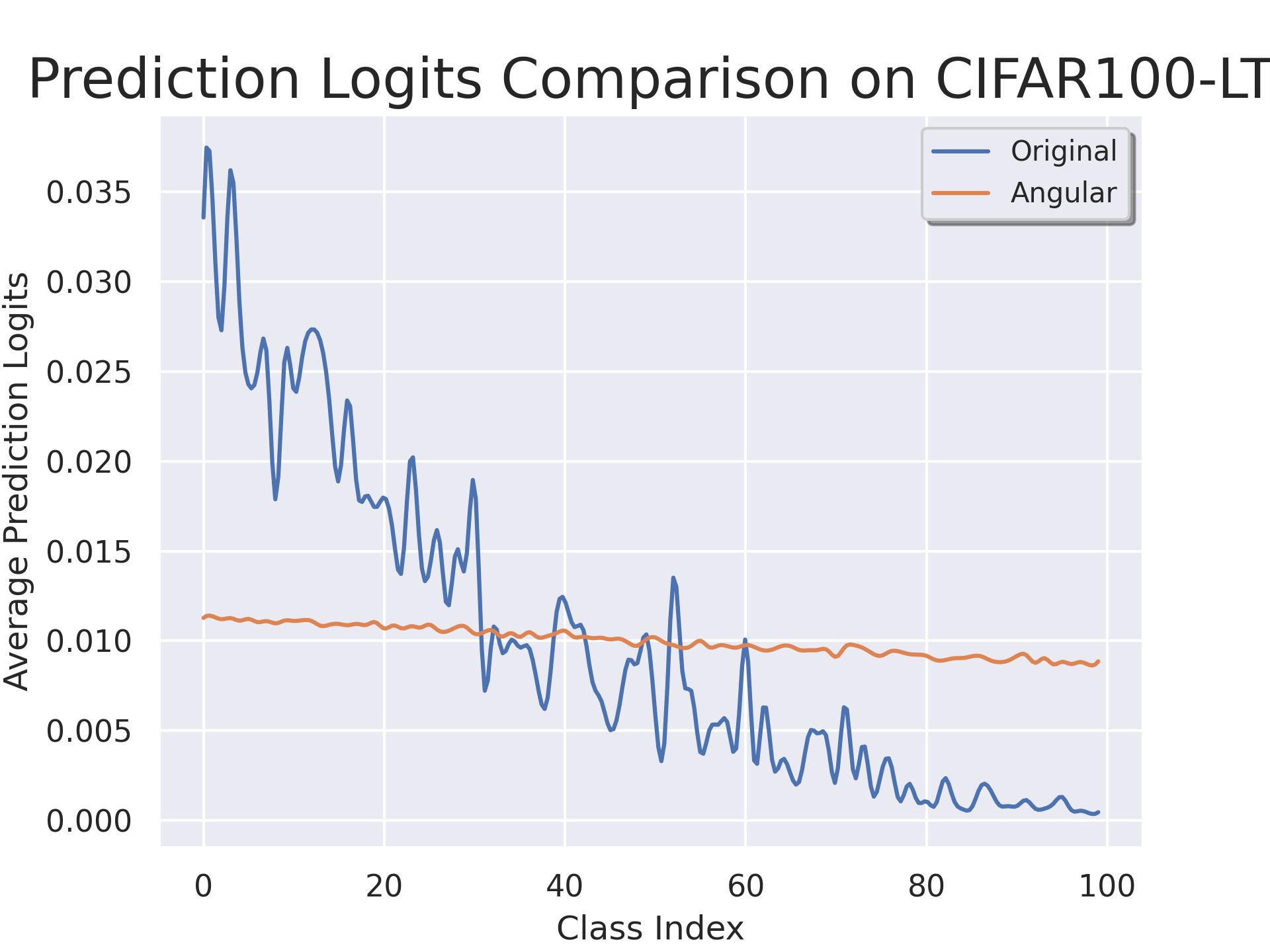}\vspace{-5pt}
   \caption{Comparison of average prediction logits on CIFAR10-LT and CIFAR100-LT validation set. Results are normalized for better visualization.}
   \label{fig:appendix_smooth}
   \vspace{-5pt}
\end{figure}

The smoothness of the prediction logits on the training set is ploted in Fig. \ref{fig:smooth_compare}, we further look into the performance in the validation set, which cannot be accessed during training. The results are shown in Fig. \ref{fig:appendix_smooth}. We can see that on the balanced validation set, the model prediction logits preserve a similar smoothness.

\subsection{Implicit Bias}
\begin{figure}[h]
  \centering
  \includegraphics[width=0.48\linewidth]{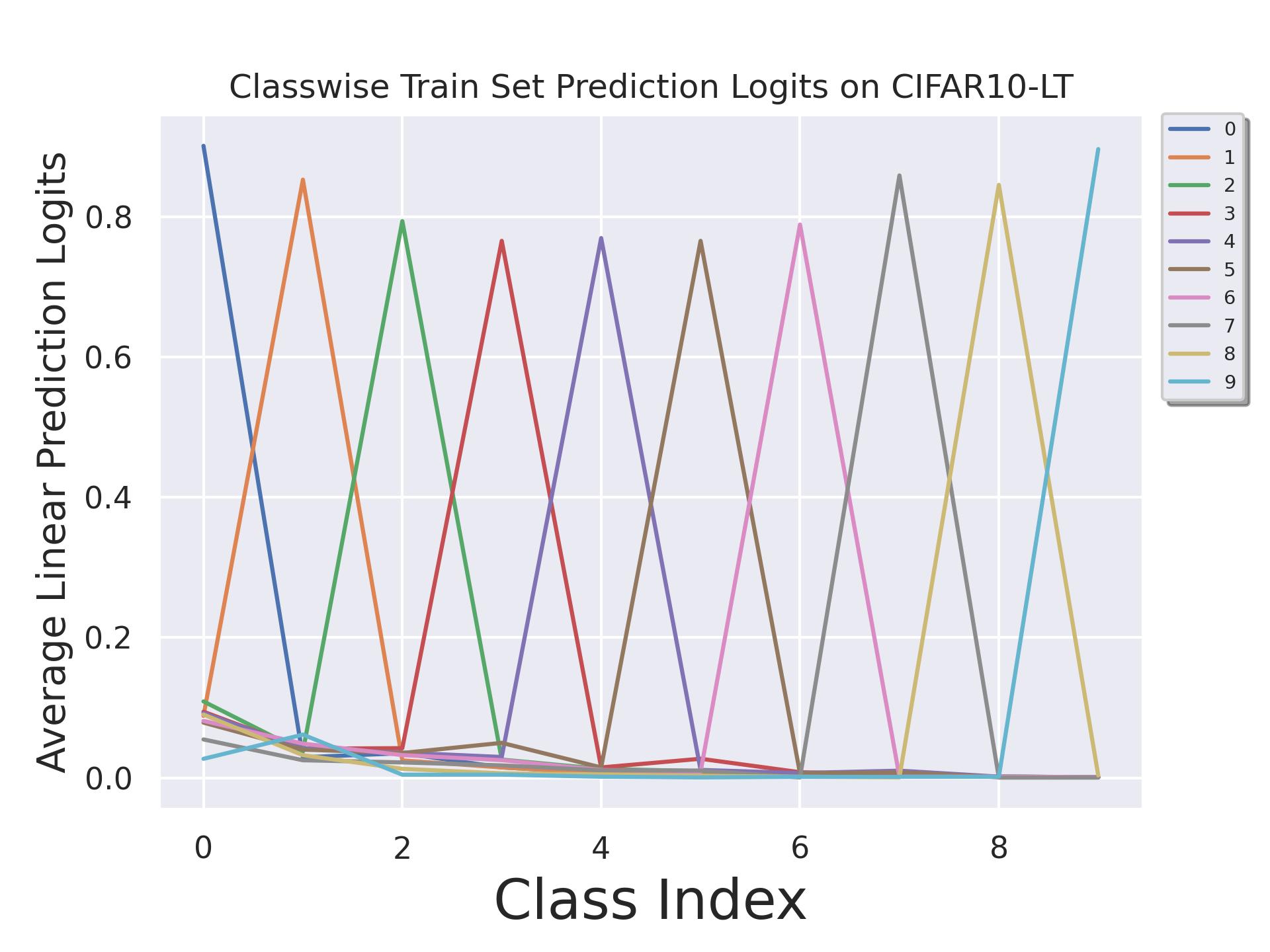}
   \includegraphics[width=0.48\linewidth]{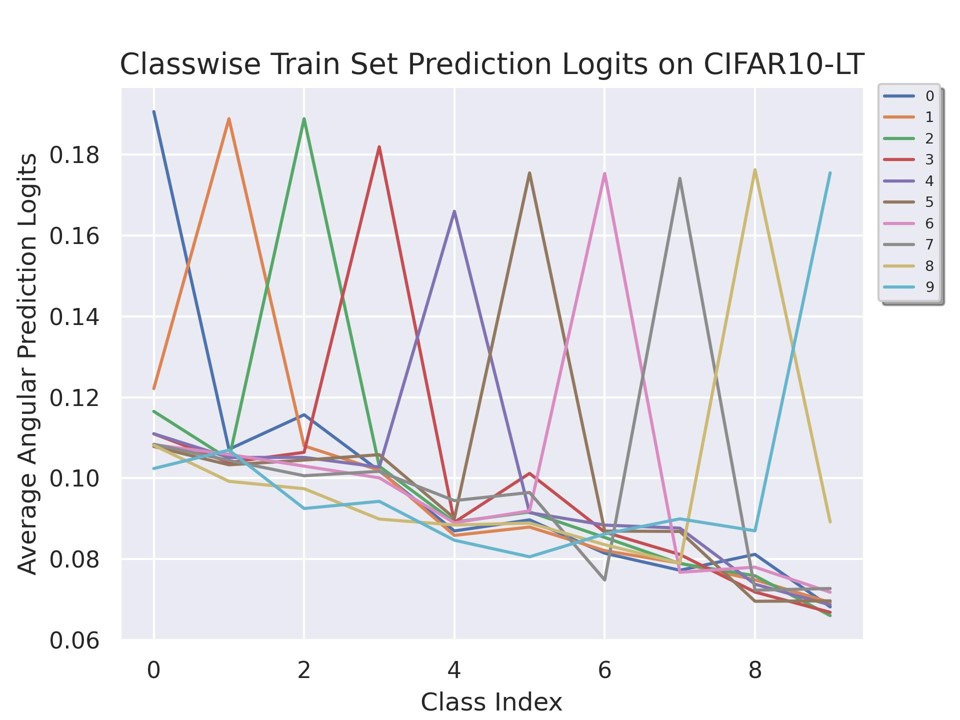}\vspace{-5pt}
   \caption{Comparison of classwise logits on CIAFR10-LT.}
   \label{fig:appendix_cifar10_classwise}
   \vspace{-5pt}
\end{figure}

\begin{figure}[h]
  \centering
  \includegraphics[width=0.48\linewidth]{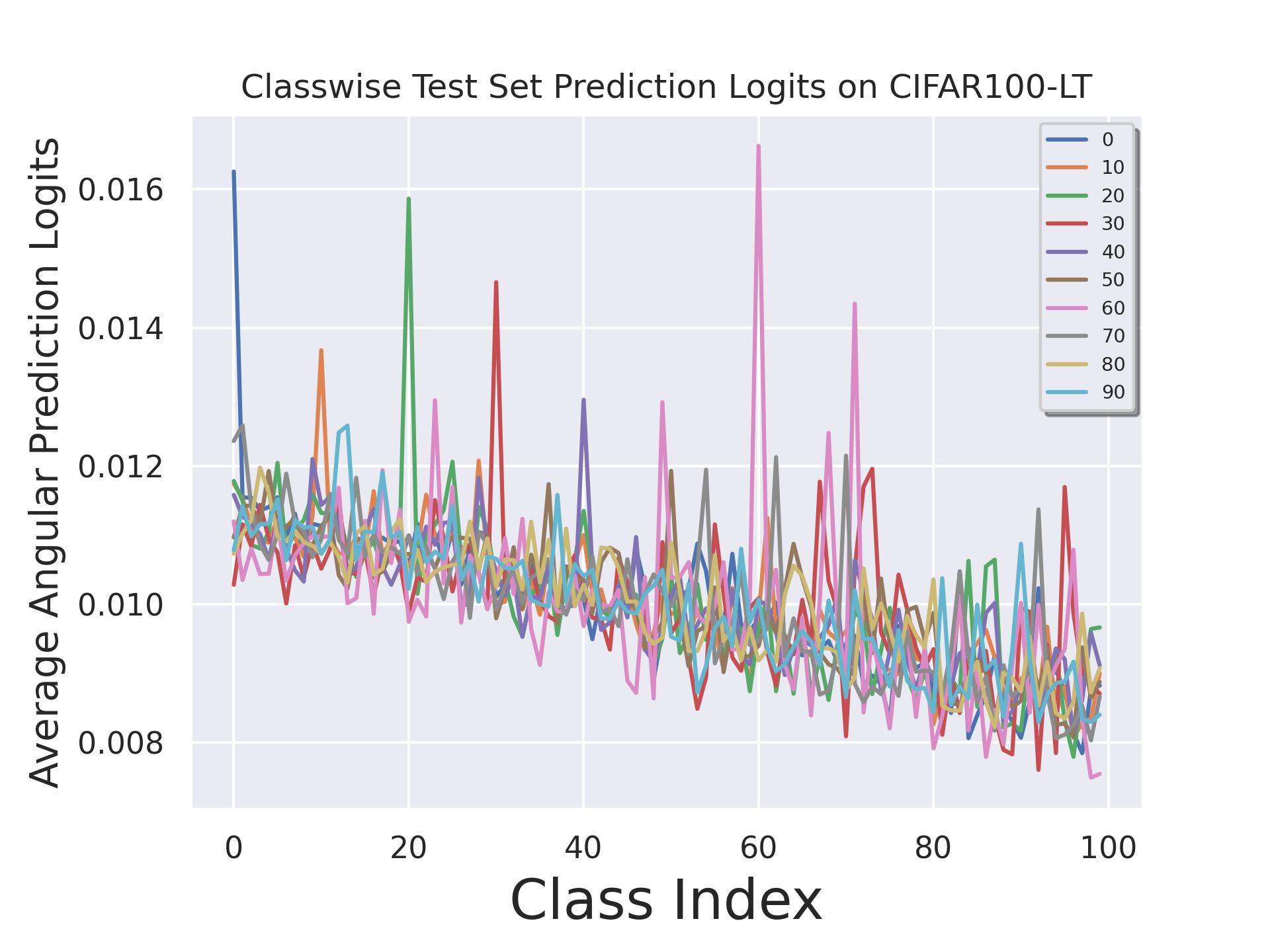}
  \includegraphics[width=0.48\linewidth]{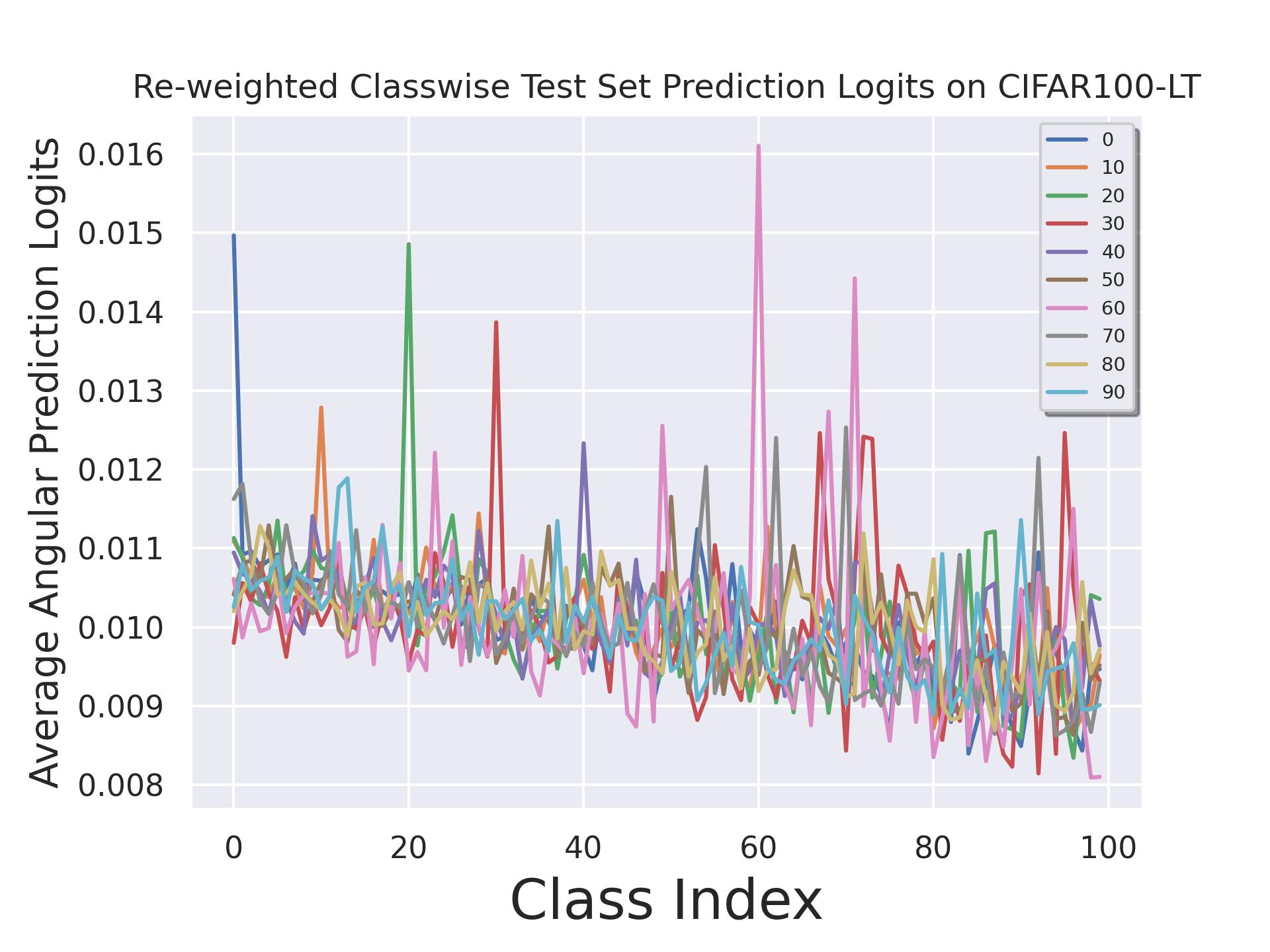} \\
  (a) Existent implicit bias \qquad  (b) Elimination by ABS \\
   \includegraphics[width=0.48\linewidth]{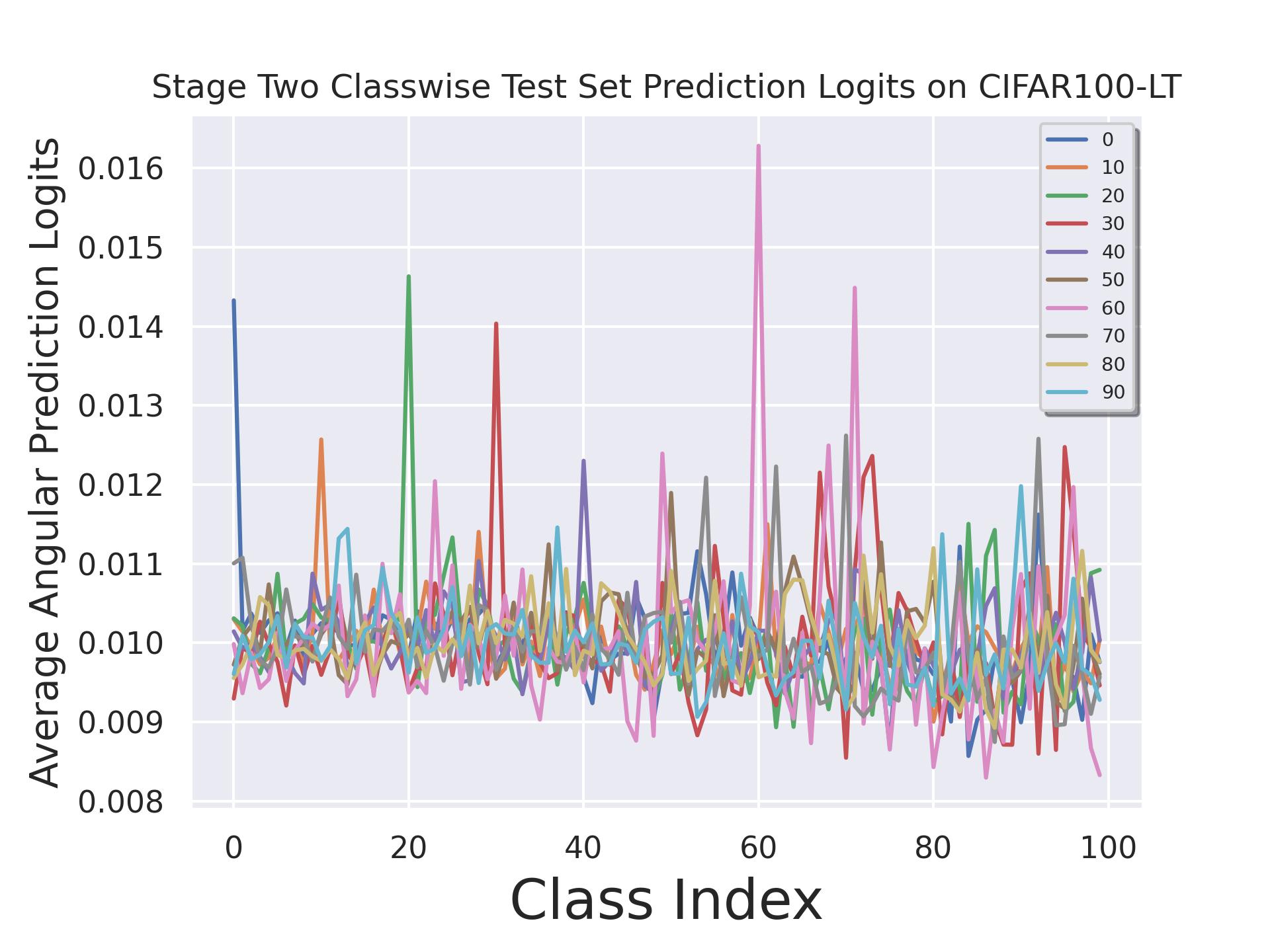} \\
   (c) Elimination by MiSLAS \\
   \vspace{-5pt}
   \caption{Comparison of implicit bias comparison on CIAFR100-LT using different baselines.}
   \label{fig:appendix_cifar100_bias_compare}
   \vspace{-5pt}
\end{figure}

As mentioned in the main paper, we will show the class-wise prediction logits comparison on CIFAR10-LT in Fig.~\ref{fig:appendix_cifar10_classwise} and the implicit bias results on CIFAR100-LT in Fig.~\ref{fig:appendix_cifar100_bias_compare}. 

\subsection{Extension to Vision Transformer}
We showed that when changing the backbone to vision transformers, the overall performance does not improve. We guess this is due to the attention mechanism of the visual transformers, thus we analyze the empirical foundation of our work on ViTs, which is the imbalance in the classifier (head). We show that the classifier imbalance does not exist, but still having an imbalanced performance over different classes, showing that the long-tailed problem cannot be fully resolved by re-balancing the classifier, and feature quality as well as quantity is important. 

Fig.~\ref{fig:vit_weight} shows the classification layer weight distribution when using backbone as ViT-B (pretrained on ImageNet-21k). We can see that the weight imbalance does not exist, while the bias term exists an imbalance (opposite trend with CNNs). This is an interesting observation, posing limitations to some methods which are built on the weight imbalance phenomenon. Our method, instead, has an interesting property of not effecting the head class performance, but making a trade-off between middle and tail classes, being still effective in improving the tail class performance.

\begin{figure}[h]
  \centering
  \includegraphics[width=0.98\linewidth]{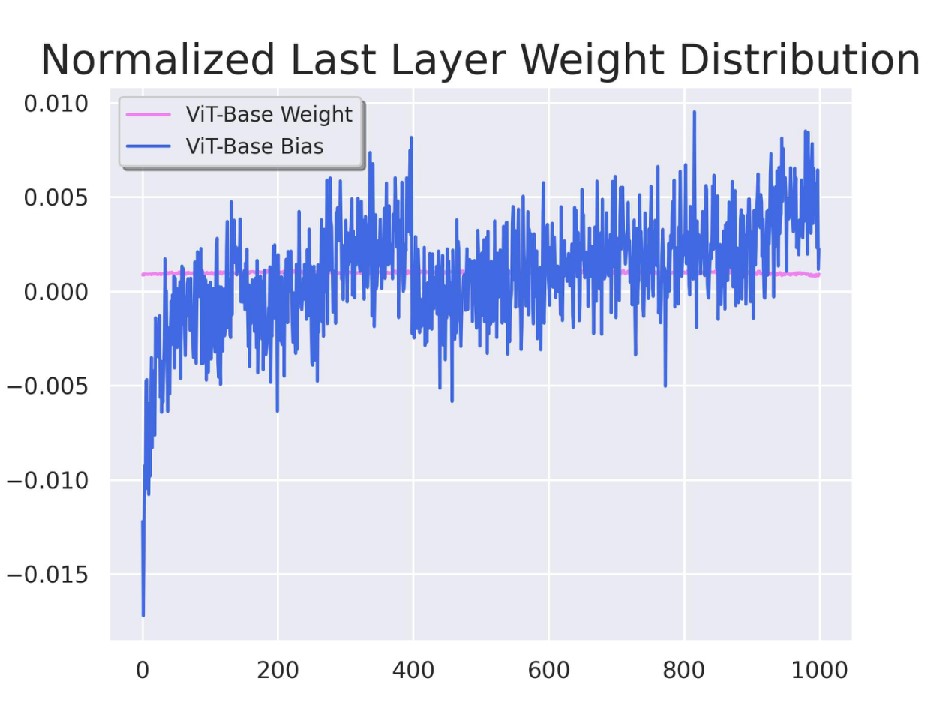}\vspace{-5pt}
   \caption{Visualization of ViT-B's classification layer weight distribution and the bias term. We show that the imbalance does not exists under the ViT backbone.}
   \label{fig:vit_weight}
   \vspace{-5pt}
\end{figure}

\section{Additional Analysis}
We further discuss some other aspects of this work: (1) We only analyzed the weight distribution of the model after one-stage training, we further discuss the effect of including stage two; (2) We empirically show that there is a natural relationship between angular information and long-tailed recognition.

\subsection{Two Stages' Weight Comparison}
\begin{figure}[h]
  \centering
  \includegraphics[width=0.48\linewidth]{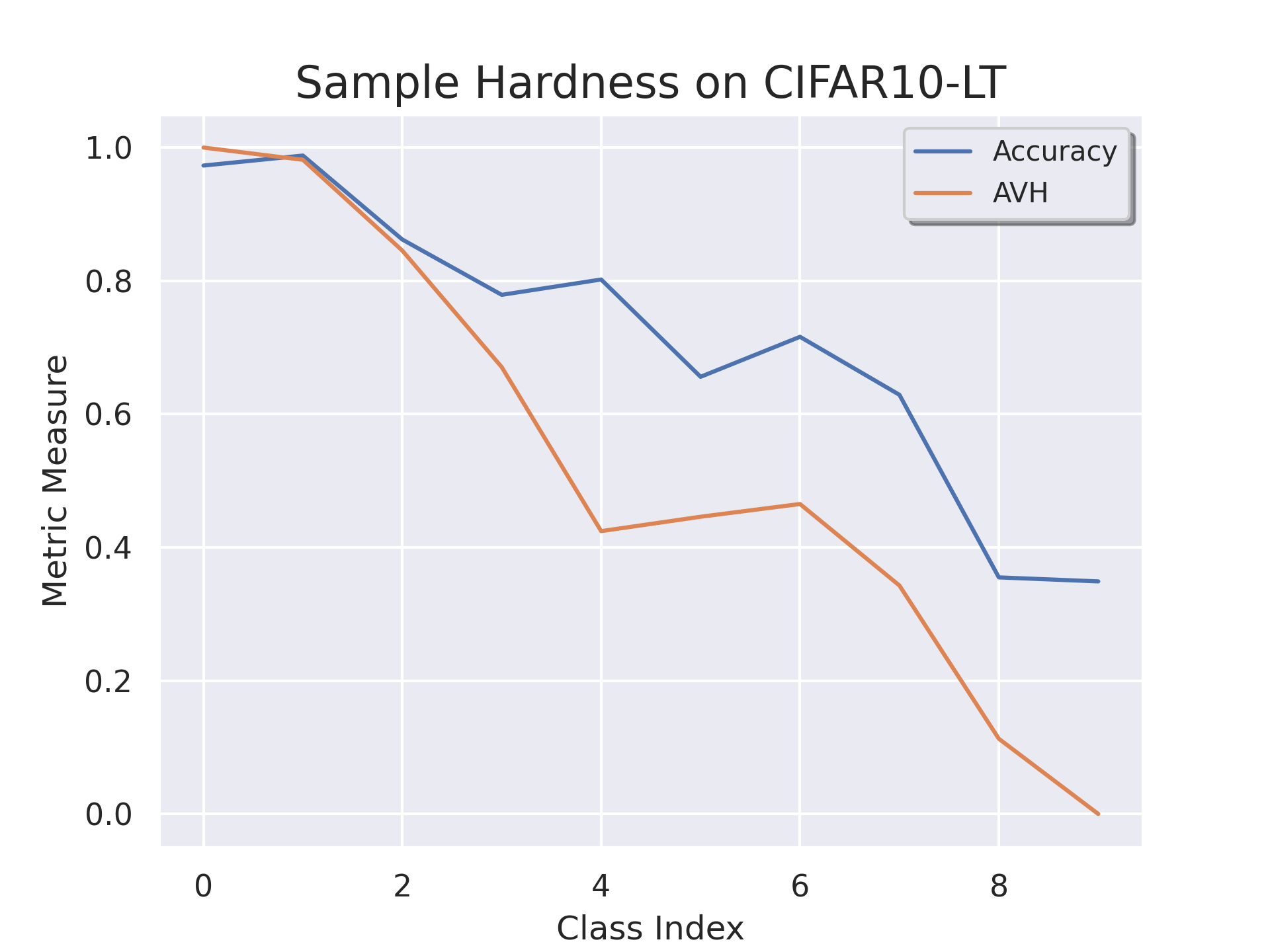}
  \includegraphics[width=0.48\linewidth]{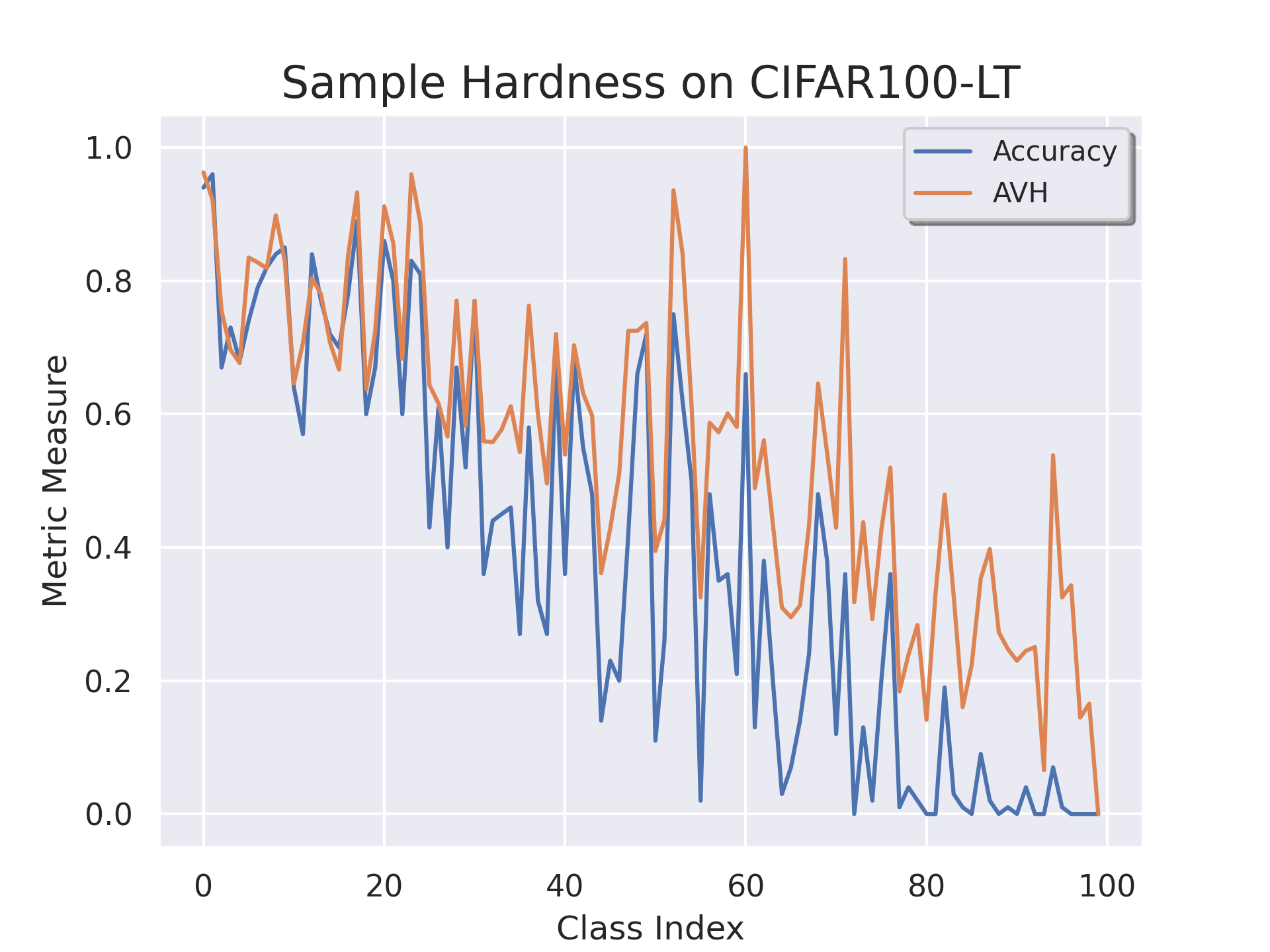}\vspace{-5pt}
   \caption{Positive correlation between performance and AVH score, indicating a natrual match between angular information and long-tailed recognition performance.}
   \label{fig:two_stage_compare}
   \vspace{-5pt}
\end{figure}

We analyze how the second stage fine-tuning effect the classifier weights, using its balanced data re-training, label-aware smoothing and linear weight scaling techniques. Fig.~\ref{fig:two_stage_compare} shows the comparison on CIFAR100-LT. We can see that it eases, but does not solves the weight imbalance. 

\subsection{Angular Hardness}
Angular information such as AVH was also proposed for measuring sample hardness. Motivated by these, we would like to verify in long-tailed recognition, whether sample hardness is correlated with the sample numbers. Fig. \ref{fig:acc_avh} shows that AVH has a positive correlation with the performance accuracy. 

\begin{figure}[h]
  \centering
  \includegraphics[width=0.48\linewidth]{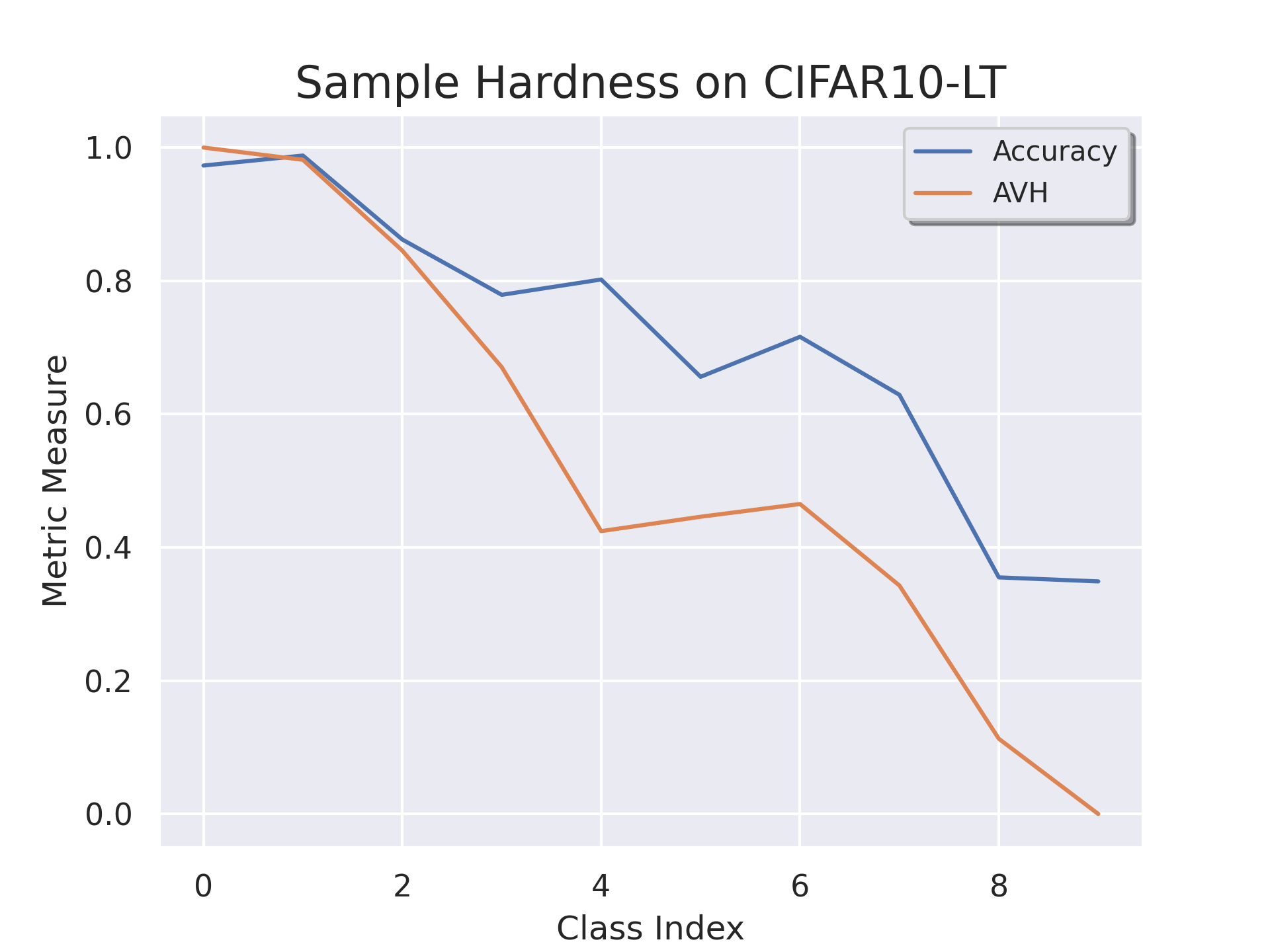}
  \includegraphics[width=0.48\linewidth]{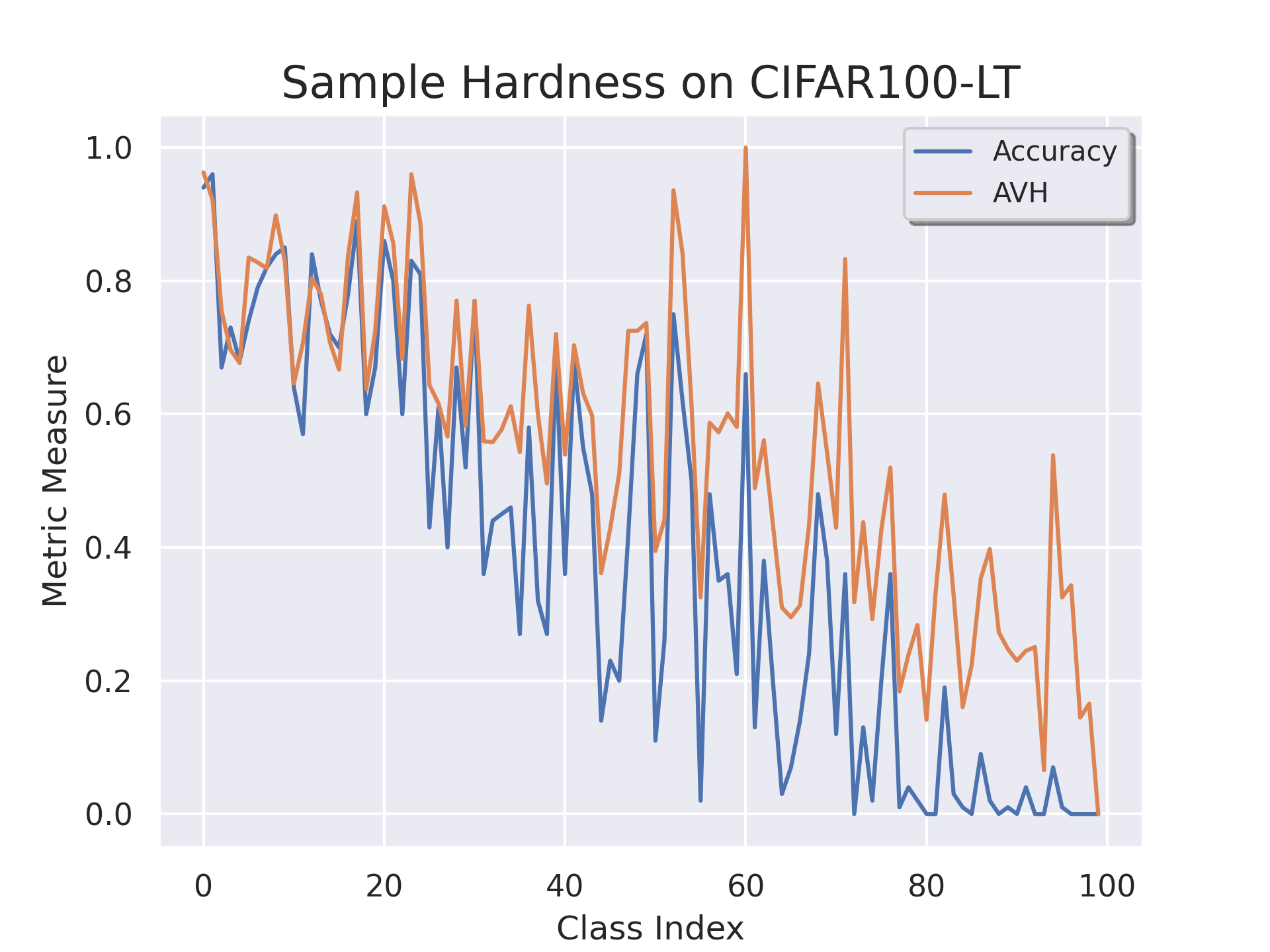}\vspace{-5pt}
   \caption{Positive correlation between performance and AVH score, indicating a natrual match between angular information and long-tailed recognition performance.}
   \label{fig:vit_weight}
   \vspace{-5pt}
\end{figure}

This makes us to interpret the angular prediction as follows: If we treat the classifier weights as different prototypes (one for each class), then the angular prediction measures the distance between the prototype and the sample. If we see the classification problem from a clustering view, then the closer a sample is to the prototype, the easier for it to be classified. Thus, it is a natural thing for us to introduce angular information into long-tailed learning, as they natrually match in terms of sample hardness.

\end{document}